\documentclass{article}

\usepackage[utf8]{inputenc} \usepackage[T1]{fontenc}    \usepackage{url}            \usepackage{booktabs}       \usepackage{amsfonts}       \usepackage{nicefrac}       \usepackage{microtype}      \usepackage{xcolor}         

\usepackage[parfill]{parskip}
\usepackage{geometry}
\usepackage{natbib}
 \usepackage[colorlinks=true,linkcolor=red,citecolor=blue]{hyperref}
 \usepackage{etoolbox}
\ifundef{\abstract}{}{\patchcmd{\abstract}{\quotation}{\quotation\noindent\ignorespaces}{}{}}

\usepackage{dsfont}
\usepackage{threeparttable}
\usepackage{amssymb}
\usepackage{mathtools}
\usepackage{float}
\usepackage{upgreek}
\allowdisplaybreaks
\usepackage{bm}
\usepackage{textcomp}
\usepackage{natbib}
\usepackage{enumitem}
\usepackage{multirow}
\usepackage{colortbl}
\usepackage{makecell}
\usepackage{microtype}
\usepackage{graphicx}
\usepackage{subfigure}
\usepackage{booktabs}
 \usepackage[linesnumbered,ruled,vlined]{algorithm2e}
\usepackage{amsthm}

\newcommand{\q}{q}

\newcommand{\p}{p}

\newcommand{\e}{E}

\newcommand{\doo}{\textnormal{do}}
\allowdisplaybreaks

\newtheorem{definition}{Definition}

\newtheorem{lemma}{Lemma}

\usepackage{eucal}

\title{\bfseries Adaptive Multi-Source Causal Inference}

\author{\normalsize\bfseries
  Thanh Vinh Vo\textsuperscript{\normalfont{ 1}}\qquad Pengfei Wei\textsuperscript{\normalfont{ 2}} \qquad Trong Nghia Hoang\textsuperscript{\normalfont{ 3}} \qquad Tze-Yun Leong\textsuperscript{\normalfont{ 1}}\\[0.3cm]
  \normalsize
  \textsuperscript{\normalfont{1}}National University of Singapore \qquad \textsuperscript{\normalfont{2}}ByteDance AI Lab\qquad \textsuperscript{\normalfont{3}}AWS AI Labs\\
}

\date{}

\begin{document}

\maketitle

\begin{abstract}
\setlength{\parindent}{0em}

Data scarcity is a tremendous challenge in causal effect estimation. In this paper, we propose to exploit additional data sources to facilitate estimating causal effects in the target population. Specifically, we leverage additional source datasets which share similar causal mechanisms with the target observations to help infer causal effects of the target population. We propose three levels of knowledge transfer, through modelling the outcomes, treatments, and confounders. To achieve consistent positive transfer, we introduce learnable parametric transfer factors to adaptively control the transfer strength, and thus achieving a fair and balanced knowledge transfer between the sources and the target. The proposed method can infer causal effects in the target population without prior knowledge of data discrepancy between the additional data sources and the target. Experiments on both synthetic and real-world datasets show the effectiveness of the proposed method as compared with recent baselines.
 \end{abstract}

\section{Introduction}
\label{sec:intro}
Estimating treatment effects of an intervention on an outcome commonly arises in many practical areas, e.g., personalized medicine \citep{henderson2016bayesian,powers2018some}, digital experiments \citep{taddy2016nonparametric} and political science \citep{green2012modeling}.
One crucial aspect in estimating treatment effect is the presence of a confounder that affects both the treatment and the outcome \citep{louizos2017causal,madras2019fairness,rakesh2018linked}. For example, a patient's socioeconomic status affects both affordable therapy options (the treatment) and the health conditions (the outcome) of this patient. In most real-world problems, the confounders cannot be directly measured, e.g., socio-economic status cannot be observed. 
Thus it can be modeled as a latent variable \citep{louizos2017causal,madras2019fairness,rakesh2018linked}.  
The key idea is then to make inference about this latent confounders using a so-called set of proxy variables (e.g., socio-economic status can be inferred by income, residential address, and so on). With the inferred confounders, the desired treatment effects can be estimated.

Inferring latent confounders usually requires sufficient data observations.
However, the observational data (i.e., proxy variables, outcomes, and treatments) of a population may be scarce in practice, possibly due to difficulty in collection or expensive annotation, leading to poor estimates of the treatment effects in that population. 
Fortunately, observations from experiments of the same treatment on different populations are likely to share similar causal mechanisms, e.g., causal graph and structural causal equations. 
This motivates us to investigate the possibility of improving the treatment effect estimation on a target population by exploiting useful information from some different but related data sources.
Following the concepts of transfer learning \citep{pan2009survey}, we refer observations for the treatment effect estimation as target population data, and the additional observations as source populations data. 
A major challenge is the presence of data divergence in different populations, e.g., the data distribution can be different across populations. For example, suppose we have sufficient observational data of a country $A$ that can be used to estimate treatment effects in that country. We wish to utilize the data of country $A$ to help infer causal effects in another country $B$ whose data is scarce. However, we find that the distribution in $B$ is different from the one in $A$. In particular, the average age and salary of people in $A$ is slightly higher than that of $B$. The gender ratio of the two countries are also different. This difference in data distribution of the two countries might contribute to the biased in causal effect estimation of country $B$ if we naively combine the two datasets. This is because the population with sufficient data observations ($A$) might dominate the one with fewer data observations ($B$), and hence leading to a bias estimation. How to adaptively transfer useful knowledge from the source $A$ to the target $B$ is a crucial problem.

\begin{figure*}
\centering
	\includegraphics[width=0.8\textwidth]{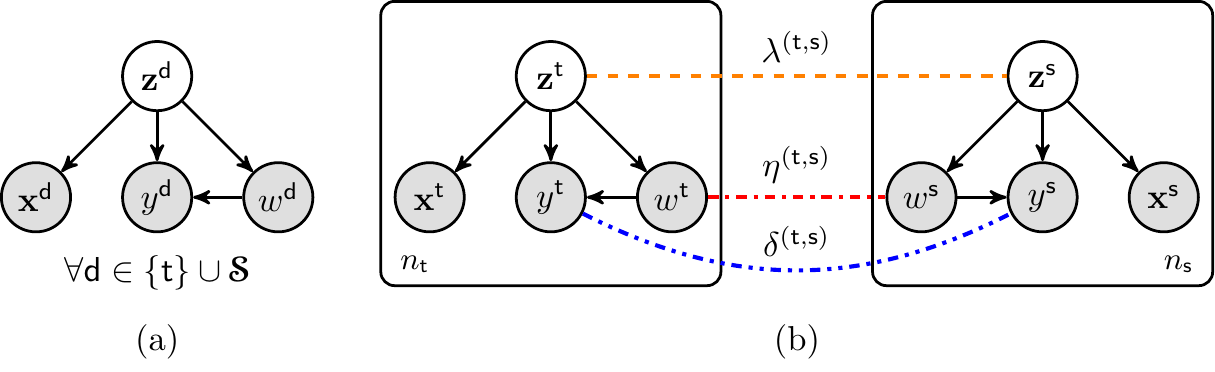}
	\vspace{-12pt}
	\caption{(a) The causal graph of population $\mathsf{d}$, where $\mathsf{d}$ can be either a source population or the target population, i.e., $\mathsf{d}\in \{\mathsf{t}\}\cup \bm{\mathcal{S}}$, where $\bm{\mathcal{S}} =\{\mathsf{s}_1,\mathsf{s}_2,\!...,\mathsf{s}_m\}$ denotes the set of $m$ sources. (b) Illustration of the three levels of knowledge transfer.  Note that the dashed lines indicate where the transfer learning happens in the inference, and they do not indicate causal relationships.} \label{fig:the-model-RT}
	\vskip -12pt
\end{figure*}

In this paper, we propose a treatment effects estimation framework that is capable of \emph{adaptively exploiting observations} from multiple sources to help infer causal effects in a target population with scarce data. The proposed method can learn the discrepancy of the source and the target population and then only transfer useful knowledge from the source populations to the target one, thus overcome the aforementioned challenges. Specifically, we focus on a casual graph as shown in Figure~\ref{fig:the-model-RT}(a). 
For the $i$--th individual, a latent confounder $\mathbf{z}_i^\mathsf{d}$ affects both the treatment $w_i^\mathsf{d}$ and the outcome $y_i^\mathsf{d}$. Herein, $\mathsf{d}$ denotes a population, which can be either a source or target population. 
The observed $\mathbf{x}_i^\mathsf{d}$ is the proxy variable that helps infer the latent $\mathbf{z}_i^\mathsf{d}$. 
We develop three levels of knowledge transfer, occurred in the inference of the outcome, treatment, and confounder. These three levels of knowledge transfer are controlled by three sets of similarity coefficients and they are learned with the observed sources and target data.  Figure~\ref{fig:the-model-RT}(b) is an illustration with one source population that helps estimate causal effects in the target population. The three similarity coefficients $\lambda^{(\mathsf{t},\mathsf{s})}$, $\nu^{(\mathsf{t},\mathsf{s})}$, $\delta^{(\mathsf{t},\mathsf{s})}$ are learned from the observed data. We name our method as Adaptive Transfer (AdaTRANS). In short, our contributions are summarized as follows:
\begin{itemize}[topsep=0pt,leftmargin=*]
\item We introduce a transfer mechanism that incorporates the target data and the additional data sources to estimate causal effects in the target population via three levels of knowledge transfer. This framework adaptively exploits the additional data sources to overcome the data scarcity issue in the target population. 

\item The advantage of the new method is that it can infer causal effects in the target population by utilizing additional data sources without prior knowledge of data discrepancy between the source populations and the target. 

\item We propose an augmented representer theorem-based variational inference procedure to approximate the posteriors of the confounding factors which leads to efficient estimation for the treatment effects.

\item Finally, we evaluate the proposed framework on an extensive benchmark comprising real-world dataset, which shows the advantage of our proposed method in comparison to the recent baselines.
\end{itemize}

\section{Related work}
\label{sec:preliminaries}
\textbf{Causal modelling without transfer.} A confounder is an important quantity in causal inference as it may induce bias in the estimates of treatment effects. To deal with the existence of confounders, the classical methods such as covariate matching, propensity score matching, Bayesian imputations for missing data \citep{rubin1974estimating,rosenbaum1983central,rubin2005causal} are based on the ignorability assumption. These formalizations are well-known as potential outcomes framework. Several modern causal effect estimators including \citet{hill2011bayesian,shalit2017estimating,hartford2017deep,Alaa:2017,yao2018representation,yoon:2018ganite,schwab2018perfect,wager2018estimation,osama2019inferring,oprescu2019orthogonal,kunzel2019metalearners,zhang2020learning,bica2020Estimating,nie2020quasi} follow the formalism of potential outcomes and these works are also based on observed confounders and the ignorability assumption. The central idea is to build a regression model of the outcomes in a causal setting. \citet{veitch2019using} propose a propensity score matching with random mini-batch data. This method also works under observed confounders and the ignorability assumption. All of these works ignore the data scarcity issue of the target population.

The following efforts take into account the unobserved confounders in causal inference: \citet{montgomery2000measuring,riegg2008causal,kuroki2014measurement,louizos2017causal,kallus2019interval,madras2019fairness,lu2018deconfounding,bica2020time,Witty2020CausalIU}.
Specifically, some proxy variables are introduced to infer the latent confounders.
For example, the household income of students is a confounder that affects the ability to afford private tuition and hence the academic performance; it may be difficult to obtain income information directly, and proxy variables such as zip code, or education level are used instead. 
These methods also work under the assumption that the observed data in the target population is sufficient for the inference of causal effects.

\textbf{Causal modelling with transfer.} A line of research that related to our work is called \emph{transportability} whose theoretical analysis has been developed in \citet{bareinboim2013transportability,bareinboim2013causal,bareinboim2014transportability,bareinboim2016causal,pearl2014external}. These works provides results on transportability of interventions on the source populations to compute causal effects on the target population. The source population can obtain experimental data by conducting randomized trials (or randomized experiments) while the target population cannot conduct such experiments. These theoretical results would also be useful for the problem of estimating causal effects from observational data only. In particular, they can be applied to observational data if the interventional distributions on the source populations can be reduced to expression with conditional distributions only using $do$-calculus of \citet{pearl1995causal}. Of interest is a recent work by \citet{aglietti2020multi}, where the source population has interventional data (experimental data), i.e., some randomised experiments are conducted on the source to collect the data. Then, a joint model of the interventional data from source population and the observational data from target population is learned. Nevertheless, there is little (or no) causal effect estimators that consider transferring causal effects while using \emph{observational data only}. In our setup, all the data that we observe has no intervention, i.e., there is no randomised experiments conducted to collect the data. Our method is in line with the theoretical results in \citet{pearl2011transportability,pearl2014external,bareinboim2013transportability,bareinboim2013causal,bareinboim2014transportability,bareinboim2016causal}. In particular, our model is a case of these theoretical works which is called \emph{trivial transportability}, where the causal effects on target population can be estimated directly with observational data of the target population only. However, due to data scarcity issue of the target population, we further develop an adaptive transfer learning algorithm to estimate causal effects in the target population using observational data of both target and source populations. The proposed method requires no prior knowledge of data discrepancy between the source and the target population since the discrepancy is learned from the observational data. Hence, it generalizes the case of naively combining  data of the source and the target population and the case of using target data only. 

Finally, we emphasize that our work using additional data sources, whose distributions may be different from that of the target data, to help infer causal effects in the target population. This problem is different from the works that utilize tools in causality to improve transfer learning algorithms such as \citet{zhang2015multi,zhang2017transfer,magliacane2018domain,rojas2018invariant,ren2018generalized,Teshima2020FewshotDA}, to name a few. %

\section{The proposed method}
This section develops a novel causal inference framework that is capable of adaptively exploiting additional data sources to help estimate treatment effects in a target population. Based on the structural causal model (SCM) \citep{pearl2000causality}, the aim of our approach is to use additional but related data sources for more accurate inference. We assume that the source observations are related with target ones in the sense that they have the identical causal graph (Figure~\ref{fig:the-model-RT}) and structural causal equations (presented in Section~\ref{sec:sce}). However, the data distributions may differ considerably across populations, which renders the failure of the straightforward data fusion\footnote{As shown in \citet{pan2009survey}, a brute-force data fusion of different populations may result in negative knowledge transfer.}. To achieve the \textit{positive} and \textit{adaptive} knowledge transfer,
we propose an augmented-representer theorem with a similarity measurement controlling the knowledge transfer strength from sources to the target population.

Specifically, we develop three levels of knowledge transfer, which take place in learning distributions of the outcome, treatment and confounder. For each level of knowledge transfer, we assign a set of learnable coefficients to model similarity of the corresponding data observations between each pair of populations, and hence adaptively transfer knowledge from multiple source populations to the target population for estimating causal effects of the target population.

\subsection{Problem description}
\label{sec:description}

In this work, we focus on one target population $\mathsf{t}$ and $m$ source populations $\mathsf{s}_1,\mathsf{s}_2,\!...,\mathsf{s}_m$. We denote the set of all source populations as $\bm{\mathcal{S}} = \{\mathsf{s}_i\}_{i=1}^m$.
For each population $\mathsf{d} \in \{\mathsf{t}\} \cup \bm{\mathcal{S}}$, we assume that a finite collection of training data tuples $\{(y_i^\mathsf{d}, w_i^\mathsf{d}, \mathbf{x}_i^\mathsf{d})\}_{i=1}^{n_\mathsf{d}}$ is provided.
Here, $w_i^\mathsf{d}$, $y_i^\mathsf{d}$ and $\mathbf{x}_i^\mathsf{d}$ denote the observed data of the treatment, the outcome and the proxy variable of individual $i$ in the population $\mathsf{d}$, respectively.
The source and target populations share the same causal graph as shown in Figure~\ref{fig:the-model-RT}, but the data distributions may be different, e.g., $\p_\mathsf{s}(\mathbf{x}_i^\mathsf{d}, w_i^\mathsf{d}, y_i^\mathsf{d}) \neq \p_\mathsf{t}(\mathbf{x}_i^\mathsf{d}, w_i^\mathsf{d}, y_i^\mathsf{d})$, where $\p_\mathsf{s}(\cdot)$ and $\p_\mathsf{t}(\cdot)$ denotes the two distributions on a source $\mathsf{s}$ and the target population $\mathsf{t}$, respectively\footnote{Herein, we only show the inequality of joint distributions. However, the marginal as well as the conditional distributions with respect to these variables can also be different.}. Moreover, the target population has scarce training observations while the source populations have sufficient training observations, $n_\mathsf{t} \ll \sum_{\mathsf{s} \in  \bm{\mathcal{S}}} n_\mathsf{s}$. 

The objective is to estimate the causal effects on a set of individuals of the target population $\mathsf{t}$ by utilizing the training observations from $\mathsf{t}$ and all source populations $\mathsf{s} \in  \bm{\mathcal{S}}$. In particular, we first train a model that utilizes training data from the source populations and the target population. Then, we use this learned model to estimate individual treatment effect (ITE) and average treatment effect (ATE) in a set of \emph{new} individuals of the target population  whose observed proxy variables are $\{\mathbf{x}_{\ast i}^\mathsf{t}\}_{i=1}^{n_\ast}$, where $n_\ast$ is the size of this set. 
These quantities are defined as follows.
\begin{definition}
\label{def:causal-effects}
Let $Y$, $W$, $X$ be random variables of the outcome, treatment, and proxy variable, respectively. Then, the individual treatment effect \emph{(ITE)} and average treatment effect \emph{(ATE)} are defined as follows
\begin{align*}
\emph{\texttt{ite}}(x) &\vcentcolon= \e\big[Y| \doo(W \!\!= \!1),X\!=\!x\big] - \e\big[Y|\doo(W \!\!=\!0),X\!=\!x\big], &\emph{\texttt{ate}} &\vcentcolon= \e[\emph{\texttt{ite}}(X)], \end{align*}
where $\doo(W\!\!=\!w)$ represents that a treatment $w \in \{0, 1\}$ is given to the individual.
\end{definition}
Note that the ITE defined here is also known as the conditional average treatment effect (CATE). From Definition~\ref{def:causal-effects}, the ITE and ATE in the above set of individuals of the target population are obtained by $\texttt{ite}(\mathbf{x}_{\ast i}^\mathsf{t})$ and $\texttt{ate} = \sum_{i=1}^{n_\ast}\texttt{ite}(\mathbf{x}_{\ast i}^\mathsf{t})/n_\ast$, respectively. 
\subsection{The procedure to estimate treatment effects}
Definition~\ref{def:causal-effects} implies that the central task to estimate ITE and ATE in the target population is to find $\e[y_i^\mathsf{t}| \doo(w_i^\mathsf{t}),\mathbf{x}_i^\mathsf{t}]$. With existence of the latent confounder $\mathbf{z}_i^\mathsf{t}$, we can further expand this quantity using the backdoor adjustment formula \citep{pearl1995causal}\footnote{A ``backdoor'' is the set of variables that is not a descendant of $w_i^\mathsf{t}$, and it blocks every path going from $y_i^\mathsf{t}$ to $w_i^\mathsf{t}$ and ending with an arrow into $w_i^\mathsf{t}$. We refer readers to \citet{pearl2009causality}, page 79, for more details.} as follows \begin{align}
\e[y_i^\mathsf{t}|\doo(w_i^\mathsf{t}), \mathbf{x}_i^\mathsf{t}] = \e_{\mathbf{z}_i^\mathsf{t}\sim \p(\mathbf{z}_i^\mathsf{t}|\mathbf{x}_i^\mathsf{t})} \big(\e[y_i^\mathsf{t}|w_i^\mathsf{t},\mathbf{z}_i^\mathsf{t}]\big).\label{eq:est-causal-eff}
\end{align}
The above equation shows that the causal effect is identifiable if we can find the conditional distributions $\p(y_i^\mathsf{t}|w_i^\mathsf{t},\mathbf{z}_i^\mathsf{t})$ and $\p(\mathbf{z}_i^\mathsf{t}|\mathbf{x}_i^\mathsf{t})$. The second distribution can be further expanded by $\p(\mathbf{z}_i^\mathsf{t}|\mathbf{x}_i^\mathsf{t}) = \int p(\mathbf{z}|\mathbf{x}_i^\mathsf{t}, y_i^\mathsf{t},w_i^\mathsf{t})  \p(y_i^\mathsf{t}|\mathbf{x}_i^\mathsf{t},w_i^\mathsf{t}) \p(w_i^\mathsf{t}|\mathbf{x}_i^\mathsf{t}) \mathrm{d}y_i^\mathsf{t}  \mathrm{d}w_i^\mathsf{t}$. 
Following the forward sampling strategy, the remaining is to find the following distributions
\begin{align}
    &\p(w_i^\mathsf{t}|\mathbf{x}_i^\mathsf{t}), &&\p(y_i^\mathsf{t}|\mathbf{x}_i^\mathsf{t},w_i^\mathsf{t}),  &&\p(\mathbf{z}_i^\mathsf{t}|\mathbf{x}_i^\mathsf{t}, y_i^\mathsf{t},w_i^\mathsf{t}), &&\p(y_i^\mathsf{t}|w_i^\mathsf{t},\mathbf{z}_i^\mathsf{t}),\label{eq:target-est}
\end{align}
and then orderly draw samples from these estimated distributions to obtain the empirical expectation of $y_i^\mathsf{t}$ given $\doo(w_i^\mathsf{t})$ and $ \mathbf{x}_i^\mathsf{t}$. Due to the data scarcity issue in the target population $\mathsf{t}$, the estimations using target observations only may not be accurate enough to recover the true ATE. 
To overcome this issue, we take into account additional data observations from all the sources $\mathsf{s} \in  \bm{\mathcal{S}}$. Consequently, we learn the distributions in Eq.~(\ref{eq:target-est}) using  training data of target population and all source populations. In the subsequent sections, we present how to \textit{adaptively} approximate these distributions.

\subsection{The structural causal equations}
\label{sec:sce}

To estimate treatment effects, we first need to specify the structural equations associated with the causal graph in Figure~\ref{fig:the-model-RT}. These equations describe causal relations among all variables in the framework. For each population $\mathsf{d} \in \{\mathsf{t}\} \cup \bm{\mathcal{S}}$, we assume the following equations.

\noindent \textbf{The latent confounder $\mathbf{z}_i^\mathsf{d}$.} 
In real world applications, it is not possible to capture all the potential confounders as some of them might not be observed due to lacking of measurement method or unknown confounders. With the existence of latent confounders, causal inference from observational data can lead to a biased estimation. The increasing availability of large and rich datasets enables unobserved confounders to be inferred from other observed variables which are known as the proxy variables. We assume the structural equation of the latent confounder as follows
 \begin{align}
     \mathbf{z}_i^\mathsf{d} = \bm{\upmu} + \bm{e}_i^\mathsf{d}, \label{eq:structural-z}
 \end{align}
 where 
 $\bm{e}_i^\mathsf{d} \sim \mathsf{N}(\mathbf{0}, \sigma_z^2\mathbf{I})$ and $\bm{\upmu}$ is a vector of $d_z$ dimensions. The choice of this structural equation is reasonable because each dimension of $\mathbf{z}_i^\mathsf{d}$ maps to a real value which gives a wide range of possible values for $\mathbf{z}_i^\mathsf{d}$. Furthermore, the Gaussian assumption of  $\bm{e}_i^\mathsf{d}$ makes it computational tractable for subsequent calculations.

\noindent \textbf{The outcome $y_i^\mathsf{d}$.} In real-life, the outcome can take different values, such as binary or a real number, depending on the the nature of data and the application. We model the outcome by the following structural equations:
\begin{align}
y_i^\mathsf{d} = \begin{dcases}
f_y\left(w_i^\mathsf{d},\mathbf{z}_i^\mathsf{d}\right) + o_i^\mathsf{d}, &\text{if } y_i^\mathsf{d}\in\mathbb{R},\\
\mathds{1}\left(o_i^\mathsf{d}\le\varphi\left(f_y\left(w_i^\mathsf{d}, \mathbf{z}_i^\mathsf{d}\right)\right)\right), &\text{if } y_i^\mathsf{d}\in\{0,1\}.
\end{dcases}
\label{eq:model-neural-y}
\end{align}
In case of continuous outcomes, $o_i^\mathsf{d} \sim \mathsf{N}(0, \sigma_y^2)$, where $\sigma_y^2$ is the variance. In case of binary outcomes, $o_i^\mathsf{d} \sim \mathsf{U}(0,1)$, where $\varphi(\cdot)$ is the logistic function and $\mathds{1}(\cdot)$ is the indicator function.  In this case, Eq.~(\ref{eq:model-neural-y}) implies that $y_i^\mathsf{d}$ given $w_i^\mathsf{d}, \mathbf{z}_i^\mathsf{d}$ follows Bernoulli distribution with $\varphi\left(f_y\left(w_i^\mathsf{d}, \mathbf{z}_i^\mathsf{d}\right)\right)$ is the probability that $y_i^\mathsf{d}=1$.
For both cases, the function $f_y(\cdot)$ is modelled in the following form:
\begin{eqnarray} \nonumber
f_y(w_i^\mathsf{d},\mathbf{z}_i^\mathsf{d}) = w_i^\mathsf{d} f_{y_1}(\mathbf{z}_i^\mathsf{d}) + (1-w_i^\mathsf{d}) f_{y_0}(\mathbf{z}_i^\mathsf{d}),
\end{eqnarray}
where $f_{y_1}\colon\mathcal{Z}\mapsto \mathcal{F}_{y_1}$ and $f^\mathsf{d}_{y_0}\colon\mathcal{Z}\mapsto \mathcal{F}_{y_0}$ are functions modelling the outcome when $w_i^\mathsf{d} = 1$ and $w_i^\mathsf{d}=0$, respectively. $\mathcal{Z}$ is the set containing $\mathbf{z}_i^\mathsf{d}$ and $\mathcal{F}_{y_1}$, $\mathcal{F}_{y_0}$ are Hilbert spaces. Since the outcome $y_i^\mathsf{d}$ is a real number, we have that $\mathcal{F}_{y_1} = \mathcal{F}_{y_0} = \mathbb{R}$. However, we will use the notation $\mathcal{F}_{y_1}$, $\mathcal{F}_{y_0}$ for convenience in presentation of the subsequent sections.

From Eqs.~(\ref{eq:est-causal-eff}) and (\ref{eq:model-neural-y}), we have that 
\begin{align}
    \e[y_i^\mathsf{t}|\doo(w_i^\mathsf{t}), \mathbf{x}_i^\mathsf{t}] =\begin{dcases}
    \e_{\mathbf{z}_i^\mathsf{t}\sim \p(\mathbf{z}_i^\mathsf{t}|\mathbf{x}_i^\mathsf{t})} \big[f_y(w_i^\mathsf{t},\mathbf{z}_i^\mathsf{t})\big], &\text{ if } y_i^\mathsf{t} \in \mathbb{R}, \\
    \e_{\mathbf{z}_i^\mathsf{t}\sim \p(\mathbf{z}_i^\mathsf{t}|\mathbf{x}_i^\mathsf{t})} \big[\varphi(f_y(w_i^\mathsf{t},\mathbf{z}_i^\mathsf{t}))\big], &\text{ if } y_i^\mathsf{t} \in \{0,1\}.
    \end{dcases}
\end{align}
This implies that the variance of $y_i^\mathsf{t}$ (e.g., $\sigma_y$) has no role in the prediction of causal effects. Thus, learning $\sigma_y$ is unnecessary. 

\noindent \textbf{The treatment $w_i^\mathsf{d}$.} Similar to the above binary case of the outcome, since $w_i^\mathsf{d}$ is also binary-valued, we specify its structural equation as follows
\begin{eqnarray}
w_i^\mathsf{d} = \mathds{1}\left( u_i^\mathsf{d}\le\varphi\left(f_{w}(\mathbf{z}_i^\mathsf{d})\right)\right),\label{eq:model-neural-w}
\end{eqnarray}
where $f_w\colon \mathcal{Z} \mapsto \mathcal{F}_w$ is a function, $\mathcal{F}_w$ is a Hilbert space, and the variable $u_i^\mathsf{d} \sim \mathsf{U}(0,1)$.

\noindent \textbf{The covariate $\mathbf{x}_i^\mathsf{d}$.} Based on values of the corresponding feature, we model its structural equation differently.
In particular, if dimension $k$ is a continuous feature, we assume that 
\begin{align}
    x_{ik}^\mathsf{d} = \begin{dcases}
    f_x (\mathbf{z}_i^\mathsf{d})_k + r_{ik}^\mathsf{d}, &\text{if } x_{ik}^\mathsf{d}\in\mathbb{R},\\
    \mathds{1}\left( r_{ik}^\mathsf{d}\le\varphi(f_x (\mathbf{z}_i^\mathsf{d})_k)\right), &\text{if } x_{ik}^\mathsf{d}\in\{0,1\},
    \end{dcases}
    \label{eq:structural-xij}
\end{align}
where the notation $f_x (\mathbf{z}_i^\mathsf{d})_k$ denotes the $k$-th dimension of $f_x (\mathbf{z}_i^\mathsf{d})$, and $r_{ik}^\mathsf{d} \sim \mathsf{N}(0,1)$ if $x_{ik}^\mathsf{d} \in \mathbb{R}$, $r_{ik}^\mathsf{d} \sim \mathsf{U}[0,1]$ if $x_{ik}^\mathsf{d} \in \{0,1\}$.

\subsection{AdaTRANS: Adaptive transfer causal inference}
\label{sec:model}

To incorporate information of source observations into the estimation of target predictive distributions in Eq.~(\ref{eq:target-est}), we utilize all of them to contribute to the learning of parameters for the target predictive distributions.
However, due to the data distribution discrepancy across populations, the contribution of the source populations needs to be regulated by the similarity of the source and target population.
That is to say, if the source and target populations are very similar, we tend to fully utilize the source observations, otherwise we should discard source ones to avoid negative transfer. 
To do so, we propose three levels of knowledge transfer via three sets of learnable similarity coefficients, occurred in the modelling of $\p(w_i^\mathsf{d}|\mathbf{x}_i^\mathsf{d})$, $\p(y_i^\mathsf{d}|\mathbf{x}_i^\mathsf{d},w_i^\mathsf{d})$, $\p(\mathbf{z}_i^\mathsf{d}|\mathbf{x}_i^\mathsf{d}, y_i^\mathsf{d}, w_i^\mathsf{d})$ and $\p(y_i^\mathsf{t}|\mathbf{z}_i^\mathsf{t}, w_i^\mathsf{t})$. We name our method as Adaptive Transfer (AdaTRANS).

\subsubsection{1\texorpdfstring{$^{st}$}{-st} transfer level: learning \texorpdfstring{$\p(\mathbf{z}_i^\mathsf{t}|\mathbf{x}_i^\mathsf{t}, w_i^\mathsf{t},y_i^\mathsf{t})$}{p(z|x,w,y)} and \texorpdfstring{$\p(y_i^\mathsf{t}|\mathbf{z}_i^\mathsf{t}, w_i^\mathsf{t})$}{p(y|z,w)}}
\label{sec:approx1}
We start with the predictive distributions that include the latent confounders. These distributions are obtained by maximizing log marginal likelihood of the joint data observations from both source and target population. Since exact inference is intractable because of the existent of latent confounders, we resort by maximize evidence lower bound (ELBO) of the marginal likelihood. 
\begin{align} 
\log\p(\mathbf{y},\mathbf{x},\mathbf{w}) &\ge \sum_{\mathsf{d}}\big(\e_{\mathbf{z}^\mathsf{d}\sim\q(\mathbf{z}^\mathsf{d}|\cdot)}\big[\log \left(\p(\mathbf{y}^{\mathsf{d}}|\mathbf{w}^{\mathsf{d}},\mathbf{z}^{\mathsf{d}})\p(\mathbf{w}^{\mathsf{d}}|\mathbf{z}^{\mathsf{d}})\p(\mathbf{x}^{\mathsf{d}}|\mathbf{z}^{\mathsf{d}})\right)\big] \nonumber\\
&\quad- {D}_{\textnormal{KL}}\big[\q(\mathbf{z}^\mathsf{d}|\cdot)\|\p(\mathbf{z}^\mathsf{d})\big]\big) =\vcentcolon \mathcal{L},\label{elbo}
\end{align}
where $\mathsf{d} \in \{\mathsf{t}\} \cup \bm{\mathcal{S}}$ denotes a population (a source or target population). We use bold-face notation $\mathbf{y}^\mathsf{d} = [y_1^\mathsf{d},\!...,y_{n_d}^\mathsf{d}]^\top$ to denote the vector of all training outcomes in population $\mathsf{d}$, and similar for the covariates $\mathbf{x}^\mathsf{d}$, treatments $\mathbf{w}^\mathsf{d}$ and latent confounders $\mathbf{z}^\mathsf{d}$. The notation $\q(\mathbf{z}_i^\mathsf{d}|\cdot)=\q(\mathbf{z}_i^\mathsf{d}|\mathbf{x}_i^\mathsf{d},w_i^\mathsf{d},y_i^\mathsf{d})$ denotes the variational posterior distribution. To be computational tractable, we use mean-field approximation, i.e., $\q(\mathbf{z}|\cdot) = \prod_d \prod_i \q(\mathbf{z}_i^\mathsf{d}|\mathbf{x}_i^\mathsf{d},w_i^\mathsf{d},y_i^\mathsf{d})$, and set the variational posterior to be normal distribution as follows
\begin{align*}
    &\q(\mathbf{z}_i^\mathsf{d}|\mathbf{x}_i^\mathsf{d},w_i^\mathsf{d},y_i^\mathsf{d}) = \mathsf{N}(\mathbf{z}_i^\mathsf{d};f_q(\mathbf{x}_i^\mathsf{d},w_i^\mathsf{d},y_i^\mathsf{d}), \sigma_q^2),\\
    &f_q(\mathbf{x}_i^\mathsf{d},w_i^\mathsf{d},y_i^\mathsf{d}) = w_i^\mathsf{d}f_{q_1}(x_i^\mathsf{d}, y_i^\mathsf{d}) + (1-w_i^\mathsf{d})f_{q_0}(x_i^\mathsf{d}, y_i^\mathsf{d}),
\end{align*}
where $f_{q_0}\colon \mathcal{X}\times\mathcal{Y} \mapsto \mathcal{F}_{q_0}$ and $f_{q_1}\colon \mathcal{X}\times\mathcal{Y} \mapsto \mathcal{F}_{q_1}$ with $\mathcal{X}$, $\mathcal{Y}$ are the sets containing $\mathbf{x}_i^\mathsf{d}$, $w_i^\mathsf{d}$, $y_i^\mathsf{d}$, respectively, and $\mathcal{F}_{q_0}$, $\mathcal{F}_{q_1}$ are Hilbert spaces. The first component in $\mathcal{L}$ can be obtained from the structural equations in Eqs.~(\ref{eq:model-neural-y})-(\ref{eq:structural-xij}). From Eq.~(\ref{elbo}), optimizing $\mathcal{L}$ allows us to learn the distributions on the target population $\p(y_i^\mathsf{t}|\mathbf{z}_i^\mathsf{t}, w_i^\mathsf{t})$ and $\q(\mathbf{z}_i^\mathsf{t}|\mathbf{x}_i^\mathsf{t},w_i^\mathsf{t},y_i^\mathsf{t})$ which is the approximation of $\p(\mathbf{z}_i^\mathsf{t}|\mathbf{x}_i^\mathsf{t},w_i^\mathsf{t},y_i^\mathsf{t})$. Learning these distributions would lead to learning the functions $f_c$ where $c\in\{y_0,y_1,q_0, q_1, x, w\}$ and the hyperparameters $\sigma_y$, $\sigma_q$. 
In the following, we optimize $\mathcal{L}$ with an adaptive transfer learning technique where we use a set of weighting factors to re-weight knowledge transfer from one population to another one.

\textbf{Adaptive transfer learning of the ELBO.} 
To learn the aforementioned distributions, we first formalize an empirical risk  and then optimize its regularized objective function.
To do that, we first draw $L$ samples of latent confounders using this relation $\mathbf{z}_i^\mathsf{d}[l] = f_q(\mathbf{x}_i^\mathsf{d},w_i^\mathsf{d},y_i^\mathsf{d}) + \sigma_q\bm{\upepsilon}_i^\mathsf{d}[l]$, where $\bm{\upepsilon}_i^\mathsf{d}[l]$ denotes a vector of $d_z$ dimensions with each element drawn from the standard normal distribution. With this procedure, we form an augmented training dataset as follows
\begin{align*}
    \mathcal{D} = \bigcup_{\mathsf{d}\in \{\mathsf{t}\} \cup \bm{\mathcal{S}}}\,\bigcup_{i=1}^{n_\mathsf{d}}\bigcup_{l=1}^L\Big\{(y_i^\mathsf{d}, w_i^\mathsf{d}, \mathbf{x}_i^\mathsf{d}, \mathbf{z}_i^\mathsf{d}[l])\Big\}.
\end{align*}
The augmented training dataset $\mathcal{D}$ is the combined data from all populations $\mathsf{d}\in \{\mathsf{t}\} \cup \bm{\mathcal{S}}$. This dataset is then substituted into the ELBO $\mathcal{L}$ to obtain the Monte-Carlo approximation of $\mathcal{L}$ whose negative quantity is the empirical risk. With the augmented training dataset, we state the following result.

\begin{lemma}
\label{lem:objective-func-rt}
Let $\widehat{\mathcal{L}}$ be the empirical risk obtatined from the evidence lower bound $\mathcal{L}$. Let $\upkappa_c$ \emph{(}$c \in \{y_0, y_1, q_0, q_1, x, w\}$\emph{)}  be kernel functions and $\mathcal{H}_c$ their associated reproducing kernel Hilbert spaces \emph{(RKHSs)}. Consider minimizing the following objective function
\begin{align}
    J = \widehat{\mathcal{L}}(f_{y_0}, f_{y_1}, f_{q_0}, f_{q_1}, f_x, f_w) + \sum_c \gamma_c \|f_c\|_{\mathcal{H}_c}^2
\end{align}
with respect to the functions $f_c$ \emph{(}$c \in \{y_0, y_1, q_0, q_1, x, w\}$\emph{)}, where $\gamma_c \in \mathbb{R}^+$. Then, the minimizer of $J$ would have the following form of each function
\begin{align}
    f_c(\bm{\upnu}_i) = \sum_j \upkappa_c(\bm{\upnu}_i, \bm{\upnu}_j)\bm{\upalpha}_j^c, \quad c \in \{y_0, y_1, q_0, q_1, x, w\},
\end{align}
where $\bm{\upnu}_{(\cdot)}$ is the input obtained from the tuples in $\mathcal{D}$. In particular, $\bm{\upnu}_{(\cdot)} = \mathbf{z}_i^\mathsf{d}[l]$ for $c=\{y_1, y_2, x, w\}$ and $\bm{\upnu}_{(\cdot)} = (\mathbf{x}_i^\mathsf{d},y_i^\mathsf{d})$ for $c \in\{q_0, q_1\}$. The coefficients $\bm{\upalpha}_j^c$ are vectors in the Hilbert space $\mathcal{F}_c$.
\end{lemma}

By minimizing $J$ with respect to $\bm{\upalpha}_j^c$ and hyperparameters, we obtain the functions $f_c$ ($c \in \{y_0, y_1, q_0, q_1, x, w\}$). Since these functions modulate the distributions $\p(y_i^\mathsf{t}|\mathbf{z}_i^\mathsf{t}, w_i^\mathsf{t})$ and $\q(\mathbf{z}_i^\mathsf{t}|\mathbf{x}_i^\mathsf{t},w_i^\mathsf{t},y_i^\mathsf{t})$, we obtain their probability density function. 

\textbf{Transferable kernel function.} The controlling of knowledge transfer is via the kernel functions $\upkappa_c$ in Lemma~\ref{lem:objective-func-rt}. Let $\mathsf{d}_1$ and $\mathsf{d}_2$ be two populations, i.e., $\mathsf{d}_1, \mathsf{d}_2 \in \{t\}\cup \bm{\mathcal{S}}$. Let $\bm{\upnu}_i^{\mathsf{d}_{1}}$ and $\bm{\upnu}_j^{\mathsf{d}_{2}}$ be two data points obtained from two tuples of the dataset $\mathcal{D}$ ($\bm{\upnu}_i^{\mathsf{d}_{1}}$ and $\bm{\upnu}_j^{\mathsf{d}_{2}}$ can be portions of the tuple depending on the input to the kernel function $\upkappa_c$). To adaptively transfer knowledge from the source populations to the target population, we use the following kernel function
\begin{align}
    \upkappa_c(\bm{\upnu}_i^{\mathsf{d}_{1}}, \bm{\upnu}_j^{\mathsf{d}_{2}}) = \begin{dcases}
    \lambda^{(\mathsf{d}_1,\mathsf{d}_2)} k_c(\bm{\upnu}_i^{\mathsf{d}_{1}}, \bm{\upnu}_j^{\mathsf{d}_{2}}), &\text{if $\mathsf{d}_1\neq\mathsf{d_2}$},\\
     \textcolor{white}{\lambda^{(\mathsf{d}_1,\mathsf{d}_2)}}k_c(\bm{\upnu}_i^{\mathsf{d}_{1}}, \bm{\upnu}_j^{\mathsf{d}_{2}}),&\text{otherwise},
    \end{dcases}\label{eq:kernel-transfer}
\end{align}
where we use a learnable parametric coefficient $\lambda^{(\mathsf{d}_1,\mathsf{d}_2)} \in [0,1]$, which we call the transfer factor, to re-weight the similarity of the two populations $\mathsf{d}_1$ and $\mathsf{d}_2$. Since there are $m$ sources populations and one target population, we would have $m(m+1)/2$ coefficients $\lambda^{(\cdot,\cdot)}$. This is the first level of knowledge transfer in our method. 
In this work, we are interested in the transfer factors between the target population $\mathsf{t}$ and a source population $\mathsf{s}$, i.e. $\lambda^{(\mathsf{t},\mathsf{s})}$ for $\mathsf{s} \in  \bm{\mathcal{S}}$. Since this coefficient helps adaptively transfer knowledge from source populations to the target population. 
If $\lambda^{(\mathsf{t},\mathsf{s})} = 1$, which indicates the two populations are highly related, this is equivalent to simply combining the source and target data.
If $\lambda^{(\mathsf{t},\mathsf{s})} = 0$, which indicates the two populations are completed unrelated, it corresponds to learning the desired distributions with target data only. For $0<\lambda^{(\mathsf{t},\mathsf{s})}<1$, the desired distributions on target population are learned with target data and partial of the source data.

\begin{lemma}
\label{lem:convex-analysis}
Let $\bm{\upalpha}_j^{q_0}$ and $\bm{\upalpha}_j^{q_1}$ be fixed. Then, the objective function $J$ in Lemma~\emph{\ref{lem:objective-func-rt}} is convex with respect to $\bm{\upalpha}_j^c$ for all $c\in \{y_0, y_1, x, w\}$. \textnormal{(}Please refer to Appendix for the proof.\textnormal{)}
\end{lemma}

Lemma~\ref{lem:convex-analysis} implies that if $\bm{\upalpha}^{q_0}$ and $\bm{\upalpha}^{q_1}$ reach its convex hull, $J$ will reach its minimal point. This is because the non-convexity of $J$ is induced by $\bm{\upalpha}^{q_0}$ and $\bm{\upalpha}^{q_1}$. This result shows that we should try different random initialization on $\bm{\upalpha}^{q_0}$ and $\bm{\upalpha}^{q_1}$ rather than the other parameters when optimizing $J$.

\subsubsection{2\texorpdfstring{$^{nd}$}{-nd} transfer level: learning \texorpdfstring{$\p(y_i^\mathsf{t}|\mathbf{x}_i^\mathsf{t},w_i^\mathsf{t})$}{p(y|x,w)}}
\label{sec:approx2}

In the previous section, we have approximated the two distributions $\p(\mathbf{z}_i^\mathsf{t}|\mathbf{x}_i^\mathsf{t}, w_i^\mathsf{t},y_i^\mathsf{t})$ and $\p(y_i^\mathsf{t}|\mathbf{z}_i^\mathsf{t}, w_i^\mathsf{t})$. To estimate the causal effects, we would need two more distributions $\p(w_i^\mathsf{t}|\mathbf{x}_i^\mathsf{t})$ and $\p(y_i^\mathsf{t}|\mathbf{x}_i^\mathsf{t},w_i^\mathsf{t})$. This section presents learning of $\p(y_i^\mathsf{t}|\mathbf{x}_i^\mathsf{t},w_i^\mathsf{t})$. We denote its approximation as $\tilde{\p}(y_i^\mathsf{t}|\mathbf{x}_i^\mathsf{t},w_i^\mathsf{t})$. Similar to the previous section, here we also adaptively transfer knowledge from source populations to the target population in learning these distribution. As all the variables involved are observed, we learn these distributions by maximizing log-likelihood of the observed data. Specifically, we model
\begin{align}
\tilde{\p}(y_i^\mathsf{d}|\mathbf{x}_i^\mathsf{d}, w_i^\mathsf{d}) =\begin{dcases}
\mathsf{N}\big(y_i^\mathsf{d};g(\mathbf{x}_i^\mathsf{d}, w_i^\mathsf{d}), \tilde{\sigma}_y^2), &\text{if } y_i^\mathsf{d}\in\mathbb{R},\\
\mathsf{Bern}\big(y_i^\mathsf{d};\varphi(g(\mathbf{x}_i^\mathsf{d}, w_i^\mathsf{d})), &\text{if } y_i^\mathsf{d}\in\{0,1\},
\end{dcases}\label{eq:transfer-y}
\end{align}
where $\tilde{\sigma}_y^2$ is the noise variance and $\varphi(\cdot)$ is the logistic function. We model the function $g(\cdot)$ as in the following form $g(\mathbf{x}_i^\mathsf{d}, w_i^\mathsf{d}) = w_i^\mathsf{d}\,g_1(\mathbf{x}_i^\mathsf{d}) + (1-w_i^\mathsf{d})\,g_0(\mathbf{x}_i^\mathsf{d})$, 
where $g_0\colon \mathcal{X} \mapsto \mathcal{F}_{y_0}$ and $g_1\colon \mathcal{X} \mapsto \mathcal{F}_{y_1}$ are functions modelling the outcome when the treatment $w_i^\mathsf{d}=0$ and $w_i^\mathsf{d}=1$, respectively.
Herein, we inject another level of knowledge transfer. Using classical representer theorem, we obtain the regularized empirical risk as follows: $J_y = \widehat{\mathcal{L}}_y(g_0, g_1) + \gamma_{y_0} \|g_0\|_{\mathcal{V}_y}^2 + \gamma_{y_1} \|g_1\|_{\mathcal{V}_y}^2$, where $\widehat{\mathcal{L}}_y(\cdot)$ is the negative log-likelihood, $\mathcal{V}_y$ is a reproducing kernel Hilbert space associated a kernel function $\psi_y(\mathbf{x}_i^{\mathsf{d}_1}, \mathbf{x}_j^{\mathsf{d}_2})$, and $\gamma_{y_0}, \gamma_{y_1}\in \mathbb{R}^+$. Herein, $\mathsf{d}_1, \mathsf{d}_2 \in \mathsf{t} \cup \bm{\mathcal{S}}$ are two populations. So we use another set of transfer factors $\delta^{(\mathsf{d_1}, \mathsf{d}_2)} \in [0, 1]$ to re-weight the cross-population similarity. This is the second level of transfer learning in our model.

\subsubsection{3\texorpdfstring{$^{rd}$}{-rd} transfer level: learning \texorpdfstring{$\p(w_i^\mathsf{t}|\mathbf{x}_i^\mathsf{t})$}{p(w|x)}}
Finally, we denotes the approximation of $\p(w_i^\mathsf{t}|\mathbf{x}_i^\mathsf{t})$ as $\tilde{\p}(w_i^\mathsf{d}|\mathbf{x}_i^\mathsf{d})$. Since the treatment is  binary, it can be modeled by the Bernoulli distribution similar to the above case when the output $y_i^\mathsf{d}$ is binary.
Concretely, we model:
\begin{align}
\tilde{\p}(w_i^\mathsf{d}|\mathbf{x}_i^\mathsf{d}) =  \mathsf{Bern}\big(w_i^\mathsf{d};\varphi(h(\mathbf{x}_i^\mathsf{d}))\big),\label{eq:transfer-w}
\end{align}
where $h\colon \mathcal{X} \mapsto \mathcal{F}_w$. Similar to the above, the regularized empirical risk obtained from the negative log-likelihood is $J_w = \widehat{\mathcal{L}}_w(h) + \gamma_w \|h\|_{\mathcal{V}_w}^2$, where $\mathcal{V}_w$ is a reproducing kernel Hilbert space associated with a kernel function $\psi_w(\mathbf{x}_i^{\mathsf{d}_1}, \mathbf{x}_j^{\mathsf{d}_2})$. Here we use another set of transfer factors $\eta^{(\mathsf{d}_1,\mathsf{d}_2)} \in [0,1]$ (where $\mathsf{d_1}, \mathsf{d}_2 \in \{\mathsf{t}\}\cup \bm{\mathcal{S}}$) to re-weight cross-population similarity. This happens the third level of knowledge transfer.

To sum up, we interweave three levels of knowledge transfer, Eqs.~(\ref{eq:kernel-transfer})~(\ref{eq:transfer-y})~and~(\ref{eq:transfer-w}) in that order. 
This enables adaptiveness in transferring knowledge between the source and the target observations.

\section{Experiments}
\label{sec:experiment}

\noindent\textbf{Baselines and the aims of our experiments.} In this section, we first perform a set of experiments to verify the effectiveness of our proposed model (AdaTRANS) in adaptively transferring knowledge from source populations to the target population, and thus improving the estimation of the treatment effects of interest. Here we aim to illustrate the importance of our proposed adaptive transfer learning method in estimating causal effects. 
Our second analysis is to compare the proposed method against some recent baselines including BART \citep{hill2011bayesian}, CFRNet \citep{shalit2017estimating}, CEVAE \citep{louizos2017causal}, OrthoRF \citep{oprescu2019orthogonal}, SITE \citep{yao2018representation}, X-learner \citep{kunzel2019metalearners}, and R-learner \citep{nie2020quasi}. Note that all the baselines do not consider data scarcity problem in the target population. The aim of this analysis is to show the efficacy of our method when some sources of data are available.

\noindent\textbf{Comparison metrics.} We report two comparison metrics: precision in estimation of heterogeneous effects (PEHE) \cite{hill2011bayesian} defined as $\epsilon_\mathrm{PEHE} = \e[((y_1 - y_0) - (\hat{y}_1 - \hat{y}_0))^2]$ and absolute error defined as $\epsilon_\mathrm{ATE} = |\e[y_1 - y_0] - \e[\hat{y}_1 - \hat{y}_0]|$, where $y_0, y_1$ are the ground truth of outcomes from the intervention and $\hat{y}_0, \hat{y}_1$ are their estimates, to evaluate ITE and ATE, respectively. The reported numbers are the out-of-sample mean and standard error over 10 replicates of the data with different random initializations of the training algorithm.

\noindent\textbf{Tuning hyperparameters and implementations of the baselines.} The setups of neural networks in \citet{louizos2017causal} (CEVAE) and \citep{yao2018representation} (SITE) closely follow that of \citet{shalit2017estimating} (CFRNet). Thus we also use these settings in our experiments. In particular, we use fully connected networks with activation function \texttt{ELU} and use the same number of hidden nodes in each hidden layer. We fine-tune all the networks with $\{1, 2,\!..., 6\}$ hidden layers, $\{50, 100, 200\}$ number of nodes per layer, and learning rate in $\{1e\text{-}1, 1e\text{-}2, 1e\text{-}3, 1e\text{-}4\}$. We reuse the code of these methods which are available online. For implementation of BART \citep{hill2011bayesian}, we use package \texttt{BartPy} which is also available online. For X-learner \citep{kunzel2019metalearners} and R-learner \citep{nie2020quasi}, we use package \texttt{causalml} \citep{chen2020causalml}. In both methods, we use \texttt{xgboost.XGBClassifier} as learners for binary outcomes and \texttt{xgboost.XGBRegressor} as learners for continuous outcomes. 
For OrthoRF \citep{oprescu2019orthogonal}, we use package \texttt{econml} \citep*{econml}.

\subsection{Synthetic data}
\noindent\textbf{Data description.} Obtaining the ground-truth of causal inference problems is a challenging task, and thus most of recent methods utilize synthetic or semi-synthetic datasets for evaluation.
In this experiment, we generate a synthetic datasets, each comprises of data from $m$ source populations ($\bm{\mathcal{S}} = \{\mathsf{s}_1,\mathsf{s}_2,\!...,\mathsf{s}_m\}$) and one target (${\mathsf{t}}$) population. Our aim is to show that the estimated ITE and ATE on the target population is closer to the true values when utilizing knowledge transferred from the source population. For each invidual $i$ in the population $\mathsf{d} \in \{\mathsf{t}\}\cup \bm{\mathcal{S}}$, we draw the latent confounder $\mathbf{z}_i^\mathsf{d}$, the proxy variable $\mathbf{x}_i^\mathsf{d}$, the treatment $w_i^\mathsf{d}$ and the outcome $y_i^\mathsf{d}$ using the following equations
\begin{align*}
&\mathbf{z}_i^\mathbf{d} \sim \mathsf{N}(\bm{0}, \sigma_z^2\mathbf{I}_2), \qquad x_{ij}^\mathsf{d} \sim \mathsf{Bern}(\varphi(a_{0j} + (\mathbf{z}_i^{\mathsf{d}})^\top \bm{a}_{1j})), \qquad w_i^\mathsf{d} \sim \mathsf{Bern}(\varphi(b_0 + (\mathbf{z}_i^{\mathsf{d}})^\top \textcolor{blue}{\bm{b}_1^\mathsf{d}})),\\
&\qquad\quad y_i^\mathsf{d}(0) \sim \mathsf{N}(\zeta(c_0 + (\mathbf{z}_i^\mathsf{d})^\top\textcolor{blue}{\bm{c}_1^\mathsf{d}}), \sigma_y^2), \qquad y_i^\mathsf{d}(1) \sim \mathsf{N}(\zeta(d_0 + (\mathbf{z}_i^\mathsf{d})^\top\textcolor{blue}{\bm{d}_1^\mathsf{d}}), \sigma_y^2),
\end{align*}
where $\varphi(\cdot)$ is the standard logistic function, $\zeta(\cdot)$ is the softplus function. We randomly set the ground truth parameters $(\sigma_z, \sigma_y, b_0, c_0, d_0) = (\sqrt{8}, \sqrt{2}, 0.5, 0.7, 2.0)$ and draw the ground truth $a_{0j} \sim \mathsf{N}(0,2)$ and $\bm{a}_{1j} \sim \mathsf{N}(0, 2\cdot\mathbf{I}_2)$ (for $j=1,2,\!...,30$). Herein, the number of dimensions of  the latent confounder $\mathbf{z}_i^\mathsf{d}$ is $d_z=2$ and the number of proxy variables is $d_x = 30$. The parameters $\textcolor{blue}{\bm{b}_1^\mathsf{d}}$, $\textcolor{blue}{\bm{c}_1^\mathsf{d}}$, $\textcolor{blue}{\bm{d}_1^\mathsf{d}}$ on different population $\mathsf{d}$ would have different ground truth values. The aim is to simulate the difference of the data distribution in different populations, which showcases the effectiveness of our model. We will describe these three parameters in the specific analyses. From the above simulation, we obtain $y_i^\mathsf{d} = w_i^\mathsf{d}\,y_i^\mathsf{d}(1) + (1-w_i^\mathsf{d})\,y_i^\mathsf{d}(0)$, i.e., $y_i^\mathsf{d} =  y_i^\mathsf{d}(1)$ when $w_i^\mathsf{d}=1$ and $y_i^\mathsf{d} =  y_i^\mathsf{d}(0)$ when $w_i^\mathsf{d}=0$. For each individual $i$, we only keep $(y_i^\mathsf{d}, w_i^\mathsf{d}, \mathbf{x}_i^\mathsf{d})$ as the observed data. 
For each population $\mathsf{d}$ (source or target), we simulate a set of $n_\mathsf{d} = 1000$ individuals. Thus, the total number source observations is $m \times 1000$. For the target data, since this data is scarce, we only use 50 for training, 100 for validation and 850 for testing. In the subsequent sections, we present the performance analysis of the proposed method compared to the baselines on this dataset.

Here, we detail the choice of our parameters used for data generation as follows. The causal relations for the data generation is based on Eq.~(\ref{fig:the-model-RT}). The treatment $w_i^\mathsf{d}$ takes binary values and it can be interpreted as patient $i$ taking (or not taking) a specific medicine. The outcome $y_i^\mathsf{d}(1)$ can be understood as blood pressure of patient $i$ after taking the medicine (the value is scaled to a specific range and we set its ground truth as normal distribution). Similarly, the outcome $y_i^\mathsf{d}(0)$ can be understood as blood pressure of patient $i$ who did not take the medicine. The latent confounder is unknown and uninterpretable and we further assume that it follows a normal distribution. Each covariate follows a Bernoulli distribution and we only use it for the data generation purpose.

\subsubsection{The importance of adaptively causal transfer learning}
\label{sec:adaptively-trans-analysis}
\begin{figure}
\centering
    \includegraphics[width=0.93\textwidth]{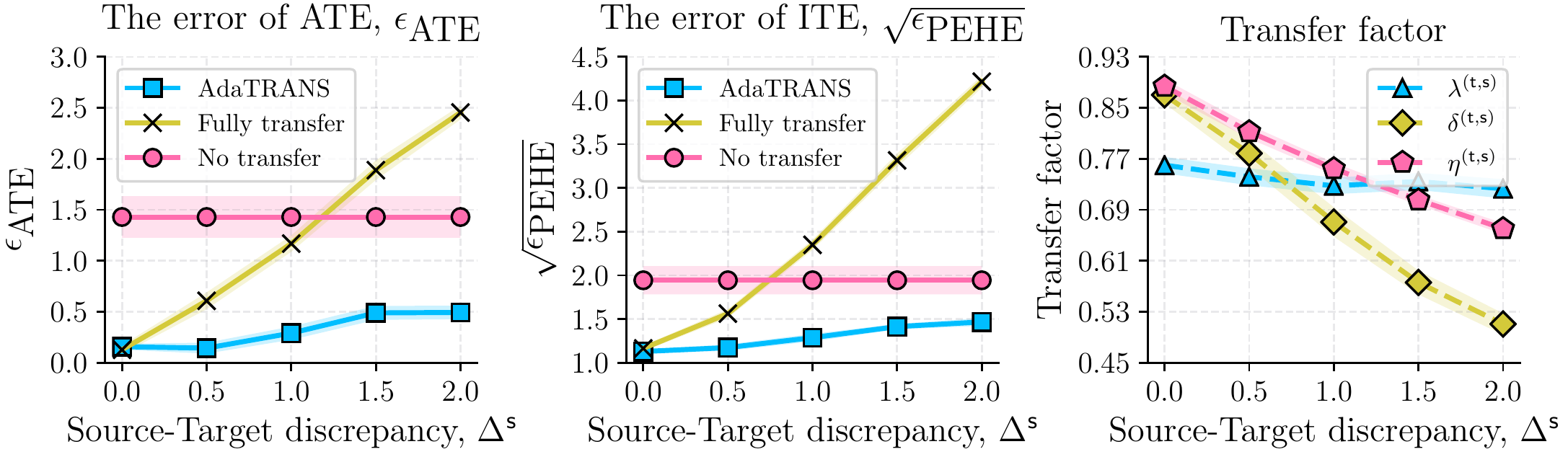}
    \caption{
       Adaptively causal transfer learning analysis.
    } \label{fig:adaptively-transfer-analysis}
\end{figure}
\begin{figure}
    \centering
    \includegraphics[width=0.73\textwidth]{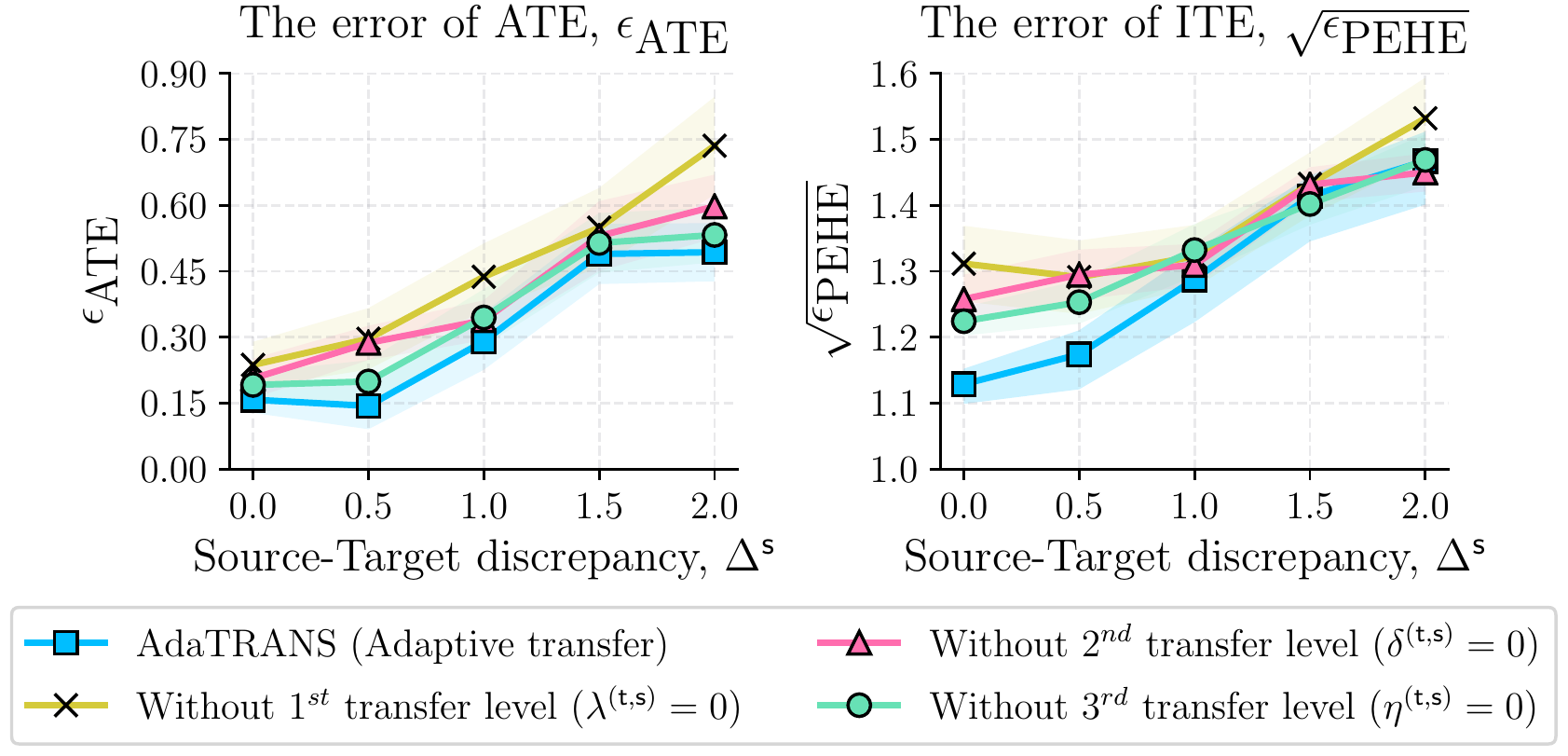}
    
    \caption{Partially causal transfer analysis.}
    \label{fig:partially-transfer-analysis}
\end{figure}
\begin{figure}
        \centering
    \includegraphics[width=0.7\textwidth]{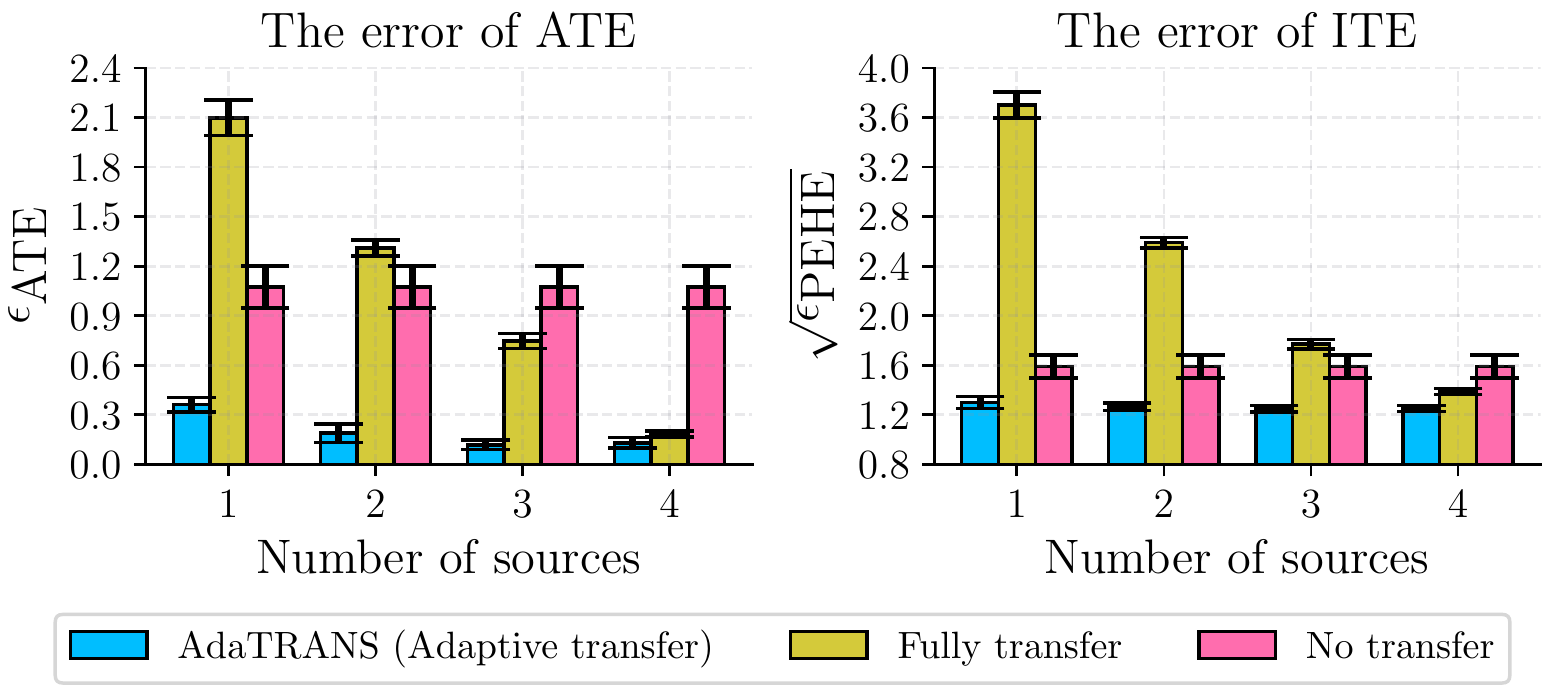}
    \caption{Multi-source causal transfer analysis.}
    \label{fig:multi-source-transfer-analysis}
\end{figure}
\noindent\textbf{Additional setups on the synthetic data.} To verify the proposed adaptively causal transfer learning model, here we use one source population $\mathsf{s}$ ($m=1$) and one target population $\mathsf{t}$. We will analyse on multi-source ($m>1$) in the subsequent sections. In this experiment, we have two sets of ground truth parameters ($\textcolor{blue}{\bm{b}_1^\mathsf{s}}$, $\textcolor{blue}{\bm{c}_1^\mathsf{s}}$, $\textcolor{blue}{\bm{d}_1^\mathsf{s}}$) and ($\textcolor{blue}{\bm{b}_1^\mathsf{t}}$, $\textcolor{blue}{\bm{c}_1^\mathsf{t}}$, $\textcolor{blue}{\bm{d}_1^\mathsf{t}}$) and we set them differently as follows
\begin{align*}
    \bm{b}_1^\mathsf{t} &= [1.1, 1.7]^\top, & \bm{c}_1^\mathsf{t} &= [1.5, 1.8]^\top, &\bm{d}_1^\mathsf{t} &= [1.5, 2.8]^\top\\
    \bm{b}_1^\mathsf{s} &= \bm{b}_1^\mathsf{t} + \Delta^\mathsf{s}[1, 1]^\top, & \bm{c}_1^\mathsf{s} &= \bm{c}_1^\mathsf{t} + \Delta^\mathsf{s}[1, 1]^\top, & \bm{d}_1^\mathsf{s} &= \bm{d}_1^\mathsf{t} + \Delta^\mathsf{s}[1, 1]^\top,
\end{align*}
where we vary $\Delta^\mathsf{s} \in \{0.0, 0.5, 1.0, 1.5, 2.0\}$ to obtain different instances of the source data. We refer to $\Delta^\mathsf{s}$ as discrepancy between the source and target population. 

\noindent\textbf{Results and discussion.} Figure~\ref{fig:adaptively-transfer-analysis} presents the performance of AdaTRANS (the proposed method, adaptively transfer, learn all transfer factors) compared to the case with fully transfer (transfer factor is set to 1) and no transfer (transfer factor is set to 0). The figure clearly shows that our propose method can adaptively learn the transfer factors $\lambda^{(\mathsf{t},\mathsf{s})}$, $\delta^{(\mathsf{t},\mathsf{s})}$, $\eta^{(\mathsf{t},\mathsf{s})}$ to control for information transfer from source to target data. In general, the more discrepant the source and target population, the lower transfer factors are. Thus, it results in better performance of our adaptively transfer than fully transfer and no transfer.

On a second analysis, we study the importance of each level of knowledge transfer in our proposed method. So we turn off one of the transferring level and observe the performance. Figure~\ref{fig:partially-transfer-analysis} illustrates the performance of each case compared to `adaptively transfer' on all levels. The figure shows that the first level (learning $\lambda^{(\mathsf{t},\mathsf{s})}$) is the most important as the performance would significantly reduce when we turn off this parameter (set $\lambda^{(\mathsf{t},\mathsf{s})}=0$). This is the transferring level of learning distributions regarding latent confounders $\mathbf{z}_i^\mathsf{t}$. Hence, learning latent confounders play a very important role in estimating causal effects.

\subsubsection{Multi-source causal transfer learning analysis}
\label{sec:multi-source-analysis}
\noindent\textbf{Additional setups on the synthetic data.} In this experiment, we simulate $m=4$ data sources where the ground truth of $\textcolor{blue}{\bm{b}_1^\mathsf{t}}$, $\textcolor{blue}{\bm{c}_1^\mathsf{t}}$, $\textcolor{blue}{\bm{d}_1^\mathsf{t}}$ are set as in the previous experiment in Section~\ref{sec:adaptively-trans-analysis}. The other parameters $\textcolor{blue}{\bm{b}_1^\mathsf{s}}$, $\textcolor{blue}{\bm{c}_1^\mathsf{s}}$, $\textcolor{blue}{\bm{d}_1^\mathsf{s}}$ (where $\mathsf{s} \in\{\mathsf{s}_1,\mathsf{s}_2,\mathsf{s}_3,\mathsf{s}_4\}$) are set as $(\Delta^{\mathbf{s}_1}, \Delta^{\mathbf{s}_2}, \Delta^{\mathbf{s}_3}, \Delta^{\mathbf{s}_4}) = (2.0, 1.5, 1.0, 0.5)$. Hence, different source populations have different levels of discrepancy to the target population.

\noindent\textbf{Results and discussion.} Figure~\ref{fig:multi-source-transfer-analysis} reports the performance of our proposed method (adaptively transfer) compared to fully transfer and no transfer. The figure shows that the more data sources, the better performance of our model. This figure again shows the superiority of `adaptively transfer' in estimating causal effects.

\begin{table}
	\caption{Out-of-sample error on synthetic dataset with different number of data sources. The dashes (\textemdash) in `1-hot' indicate that the numbers are the same as those of `stack'.}
	\vspace{3pt}
	\setlength{\tabcolsep}{2.9pt}
	\small	
	\centering
\begin{tabular}{lcccccc}
		\toprule
		\multicolumn{1}{l}{\multirow{2}{*}{Method}} & \multicolumn{3}{c}{The error of ITE ($\sqrt{\epsilon_\textrm{PEHE}}$)}       & \multicolumn{3}{c}{The error of ATE ($\epsilon_\textrm{ATE}$)}       \\ \cmidrule(lr){2-4}\cmidrule(lr){5-7}
		& 0-source & 2-sources & 4-sources & 0-source& 2-sources &4-sources \\\cmidrule(lr){1-1}\cmidrule(lr){2-4}\cmidrule(lr){5-7}
CEVAE$_{\textrm{stack}}$ \citep{louizos2017causal}  &3.13$\pm$.30   &4.55$\pm$.39              &4.81$\pm$.40 &1.70$\pm$.29  &2.77$\pm$.30 &2.48$\pm$.26    \\
CFRNet$_{\textrm{stack}}$ \citep{shalit2017estimating}    &4.55$\pm$.51  &8.87$\pm$.50   &6.01$\pm$.19 &1.64$\pm$.41 &6.09$\pm$.48 &3.98$\pm$.17                      \\
SITE$_{\textrm{stack}}$ \citep{yao2018representation}  & 5.98$\pm$.98  &8.89$\pm$.61   &7.49$\pm$.60 &3.27$\pm$.67 &6.40$\pm$.79 &5.03$\pm$.76                      \\         
		BART$_{\textrm{stack}}$ \citep{hill2011bayesian} &2.47$\pm$.06   &2.27$\pm$.03 &2.18$\pm$.06 &1.16$\pm$.13 &0.72$\pm$.08 &0.56$\pm$.09    \\
		R-learner$_{\textrm{stack}}$ \citep{nie2020quasi} &2.96$\pm$.27   &2.17$\pm$.11 &1.81$\pm$.09 &1.36$\pm$.35 &1.19$\pm$.17 &0.99$\pm$.10    \\   
		X-learner$_{\textrm{stack}}$ \citep{kunzel2019metalearners} &2.03$\pm$.13   &2.15$\pm$.12 &1.92$\pm$.13 &\textbf{1.00$\pm$.17} &1.04$\pm$.11 &1.06$\pm$.13   \\
		OrthoRF$_{\textrm{stack}}$ \citep{oprescu2019orthogonal} &6.19$\pm$.40   &2.43$\pm$.03 &2.21$\pm$.03 &1.21$\pm$.37 &0.48$\pm$.08 &0.56$\pm$.06      
       \\\cmidrule(lr){1-1}\cmidrule(lr){2-4}\cmidrule(lr){5-7}
CEVAE$_{\textrm{1-hot}}$ \citep{louizos2017causal}  &\textemdash   &5.02$\pm$.43 &3.26$\pm$.12 &\textemdash  &3.13$\pm$.42 &1.92$\pm$.23    \\
		CFRNet$_{\textrm{1-hot}}$ \citep{shalit2017estimating}    &\textemdash   &4.41$\pm$.26 &3.31$\pm$.21 &\textemdash  &3.28$\pm$.26 &2.13$\pm$.17                      \\
		SITE$_{\textrm{1-hot}}$ \citep{yao2018representation}  &\textemdash   &5.80$\pm$.99 &3.20$\pm$.25 &\textemdash  &3.41$\pm$.67 &2.13$\pm$.21                      \\         
		BART$_{\textrm{1-hot}}$ \citep{hill2011bayesian} &\textemdash   &2.26$\pm$.03 &2.18$\pm$.04 &\textemdash  &0.65$\pm$.10 &0.43$\pm$.10   \\   
		R-learner$_{\textrm{1-hot}}$ \citep{nie2020quasi} &\textemdash   &2.03$\pm$.07 &1.68$\pm$.15 &\textemdash &0.84$\pm$.15 &0.84$\pm$.20    \\   
		X-learner$_{\textrm{1-hot}}$ \citep{kunzel2019metalearners} &\textemdash   &1.90$\pm$.12 &1.82$\pm$.10 &\textemdash &0.72$\pm$.13 &0.56$\pm$.12   \\
		OrthoRF$_{\textrm{1-hot}}$ \citep{oprescu2019orthogonal} &\textemdash   &5.49$\pm$.30 &4.09$\pm$.16 &\textemdash &3.93$\pm$.22 &2.55$\pm$.17      
       \\\cmidrule(lr){1-1}\cmidrule(lr){2-4}\cmidrule(lr){5-7}
AdaTRANS  &\textbf{1.59$\pm$.09}  &\textbf{1.26$\pm$.03}  &\textbf{1.25$\pm$.02}
		&1.07$\pm$.13 &\textbf{0.19$\pm$.05} &\textbf{0.13$\pm$.03}
\\
		\bottomrule
	\end{tabular}
	\label{tab:mae}
\end{table}

\subsubsection{Performance analysis: compare with the baselines} 
\label{sec:performance-synthetic-baselines}

\noindent\textbf{Two setups of the baselines.} In this section, we compare AdaTRANS with the baselines. For each baseline, we train with two cases as follows. (1) In the first case, we combine sources and target data by adding a categorical covariate to indicate the population of each entry in the dataset. This categorical covariate is then transformed into a `1-hot vector' for training the models. (2) In the second case, we combine source and target data by `stacking' them, i.e., there is no additional covariate. The data we use in this analysis is the one simulated in Section~\ref{sec:multi-source-analysis}. 

\noindent\textbf{Results and discussion.} Table~\ref{tab:mae} reports the  performance of each method in estimating ATE and ITE. The figure shows a significant improvement when adding more data sources. The reason here is because the other baselines do not consider adaptively transfer learning, thus results in negative transfers that reduce their performance. Our model, in contrast, only transfers useful knowledge and hence achieves better performance. The figures also reveal that negative transfer happened in CEVAE, CRFNet and SITE since the performance of these method reduced when adding more data sources that are different from the target data. In addition, the baselines trained with  combination of sources and target data by adding a categorical covariate to indicate the population (1-hot) tends to give better performance than on the stacking datasets (stack).

\subsection{Twins dataset}
\begin{figure}
    \centering
		\includegraphics[width=0.7\textwidth]{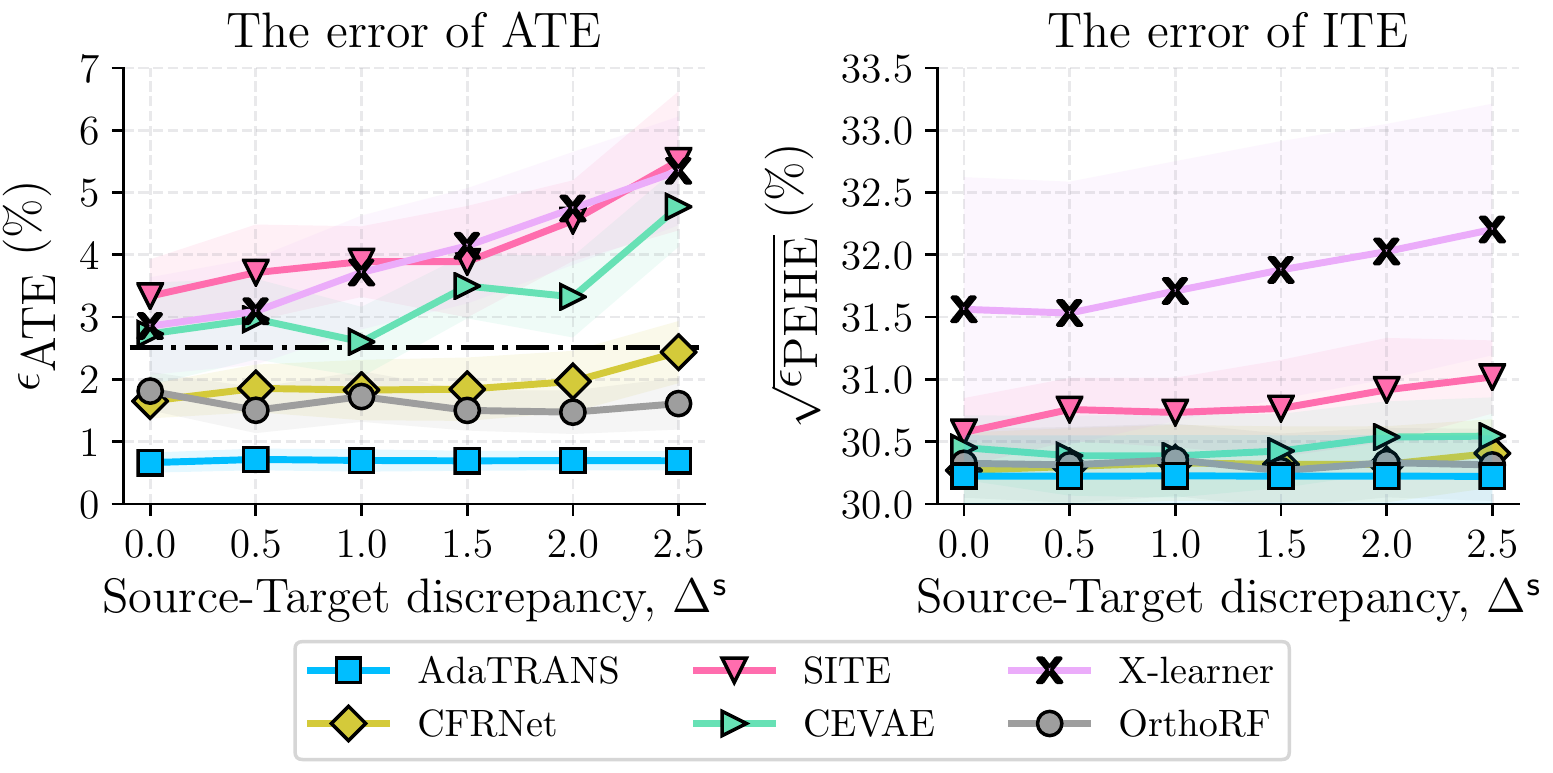}
		\caption[caption]{Out-of-sample error of ATE and ITE on Twins dataset. The dasked black line on the left figure is the error of using the naive ATE estimator: the difference between the average treated ($w_i^\mathsf{t}=1$) and average control outcomes ($w_i^\mathsf{t}=0$).}
		\label{fig:twins-transfer-factor-analysis}
\end{figure}
\noindent\textbf{Data description.} The Twins dataset contains multiple records of twin births in the US from 1989 to 1991 \citep{louizos2017causal}. In this dataset, an abstract treatment $w = 1$ corresponds to the population that gives birth to twins with heavier weight and likewise, $w = 0$ to twins born with lighter weight. The treatment outcome corresponds to the mortality of each of the twins in their first year of life. Since there are records for both twins, the mortality of twins has two possible outcomes (e.g., dead or alive) with respect to the treatment $w = 1$. Following \citet{louizos2017causal}, we focused on twins with both weighting less than 2kg. The observational study is simulated as follows. For each pair of twins, observation regarding one of them is randomly excluded. The entire dataset is then partitioned into two sets: source and target data. The source data accounts for 81\% (3921 entries) and the target data account for 19\% (900 entries). In the target data, we use 9-fold cross-validation with 100 entries for training, 100 for validation, and 700 for testing. 

\noindent\textbf{Simulation of latent confounders.} To simulate the case of latent confounders with proxy variables, the treatment assignment on twins is based on feature \texttt{GESTAT10}, which records the number of gestation weeks prior to birth and is highly correlated with the mortality outcome. We obtain the observed treatments by drawing from the following distribution $w_i^\mathsf{d}\,|\,z_i^\mathsf{d} \sim \text{Bern}(\varphi(b^\mathsf{d}(0.1z_i^\mathsf{d}-0.1)))$, where $\mathsf{d}\in \{\mathsf{s},\mathsf{t}\}$ and $z_i^\mathsf{d}$ is \texttt{GESTAT10}. We set $b^\mathsf{t} = 0.2$ and $b^\mathsf{s} = b^\mathsf{t} +\Delta^\mathsf{s}$ to simulate discrepancy between the source and target data. We vary $\Delta^\mathsf{s} \in \{0.0, 0.5, 1.0, 1.5, 2.0, 2.5\}$. Following \citet{louizos2017causal}, we created proxies $\mathbf{x}_i^\mathsf{d}$ for the hidden confounder $\mathbf{z}_i^\mathsf{d}$ as follows: The 10 categories of feature \texttt{GESTAT10} are encoded with one-hot encoding, and replicate it 3 times. We use three replications to ensure that the confounder can be recovered  \citep[follow from][]{kruskal1976more,allman2009identifiability,anandkumar2014tensor,louizos2017causal}. The resulting target data has 30 dimensions for proxy variable $\mathbf{x}_i^\mathsf{d}$.

\noindent\textbf{Results and discussion.} The performance of AdaTRANS and the baselines are presented in Figure~\ref{fig:twins-transfer-factor-analysis}. It can be observed that AdaTRANS achieves lower error at all levels of discrepancy between source and target data. This again demonstrates the advantages of our method in adaptively transferring source knowledge for target causal effect estimation. 
In Figure~\ref{fig:twins-transfer-factor-analysis}~(left), dashed dotted black line, we present error of a naive ATE estimator where ATE is estimated by simply taking difference of the average outcomes of the two groups `treated' ($w_i^\mathsf{d}=1$) and `control' ($w_i^\mathsf{d}=0$). The figure reveals that AdaTRANS, CFRNet, and OrthoRF are better than the naive estimator while the others perform worse than this naive estimator. 
Note that BART does not support binary outcomes so we do not report it on this dataset. R-learner seems to perform the worst with all of our fine-tuning setups. For a clear illustration, we report experiments on rest methods here in Figure~\ref{fig:twins-transfer-factor-analysis} and defer the experiments of R-learner to Appendix.

Here, we further discuss on why the performance in predicting ATE of the other methods are not better than that of the naive estimator. From Figure~\ref{fig:twins-transfer-factor-analysis}~(left), the performance of SITE and X-learner in predicting ATE ($\epsilon_\text{ATE}$) is low, this might be because those methods only consider observed confounders (through the unconfoundedness assumption) and ignore latent confounders. Although CEVAE can handle latent confouders, it is based on deep neural networks which needs a huge dataset to achieve a good performance. Importantly, since SITE and CEVAE are based on neural networks, their performance depends on the tuning of hyperparameters and the optimization algorithm. Hence, they might lead to a local optima. Our method, on the other hand, models complex non-linear function using kernel functions while obtaining convex or conditionally convex objective functions (as stated in Lemma~\ref{lem:convex-analysis}).

\section{Conclusion}
\label{sec:conclusion}
We have developed a knowledge transfer method to estimate the causal effects of an intervention with limited amount of observational data for a target population. To improve the estimation quality, we propose a new causal estimation method that leverages observational data from some other sources, generated from similar causal intervention mechanisms. The proposed method requires no prior knowledge on data discrepancy of the source and the target population. Experiments on both synthetic and real-world data show that the proposed method achieves competitive results to the baselines. 

A potential limitation of our approach lies in scope of our work where we assume that the sources and target population share the same causal graphs and the causal effects in all populations are identifiable with their own observed data. A practical interest for future research is to generalize the problem where causal effects in the target population may be unidentifiable. In such situation, utilizing additional data sources to help infer causal effects in the target population would be an interesting problem that need further efforts to tackle.

\bibliographystyle{apalike}
\bibliography{ref}

\appendix
\newpage
\begin{center}
    \centering \LARGE\bfseries Appendix: \\Adaptive Multi-Source Causal Inference
\end{center}
\vspace{1.5cm}
\section{Additional experimental results}
This section presents experiments on an additional dataset IHDP and additional results on the synthetic dataset, as well as the Twins dataset.

\subsection{Additional dataset: IHDP}

\begin{table}[!ht]
	\caption{Out-of-sample error on IHDP dataset with different number of data sources. The dashes (\textemdash) in `1-hot' indicate that the numbers are the same as those of `stack'.}
	\vspace{3pt}
	\setlength{\tabcolsep}{5pt}
\centering
\begin{tabular}{@{}lcccc@{}}
		\toprule
		\multicolumn{1}{l}{\multirow{2}{*}{Method}} & \multicolumn{2}{c}{The error of ITE ($\sqrt{\epsilon_\textrm{PEHE}}$)}       & \multicolumn{2}{c}{The error of ATE ($\epsilon_\textrm{ATE}$)}       \\ \cmidrule(lr){2-3}\cmidrule(lr){4-5}
		& 0-source & 1-sources & 0-source& 1-sources  \\\cmidrule(lr){1-1}\cmidrule(lr){2-3}\cmidrule(lr){4-5}
CEVAE$_{\textrm{stack}}$ 
&4.38$\pm$2.11   &4.09$\pm$2.01 &2.39$\pm$1.06 &1.68$\pm$0.79    \\
CFRNet$_{\textrm{stack}}$ 
&5.62$\pm$2.60   &5.54$\pm$2.66  &4.15$\pm$1.77 &4.06$\pm$1.84                      \\
SITE$_{\textrm{stack}}$ 
&5.84$\pm$2.76   &5.90$\pm$2.74 &4.45$\pm$1.98 &4.57$\pm$1.96                      \\         
		BART$_{\textrm{stack}}$ 
&5.44$\pm$2.68   &4.37$\pm$2.29 &3.85$\pm$1.88 &2.25$\pm$1.26    \\
		R-learner$_{\textrm{stack}}$ 
&5.47$\pm$2.49   &2.93$\pm$1.12 &3.05$\pm$1.88 &\textbf{0.70$\pm$0.42}   \\   
		X-learner$_{\textrm{stack}}$ 
&3.90$\pm$2.06   &2.64$\pm$1.09 &2.48$\pm$1.61 &1.00$\pm$0.49   \\
		OrthoRF$_{\textrm{stack}}$ 
&4.91$\pm$2.38   &2.97$\pm$1.65 &3.10$\pm$1.76 &2.01$\pm$1.22      
      \\\cmidrule(lr){1-1}\cmidrule(lr){2-3}\cmidrule(lr){4-5}
CEVAE$_{\textrm{1-hot}}$ 
&\textemdash   &4.16$\pm$2.07 &\textemdash &1.91$\pm$0.88   \\
		CFRNet$_{\textrm{1-hot}}$ 
&\textemdash   &5.54$\pm$2.66 &\textemdash &4.05$\pm$1.84   \\
		SITE$_{\textrm{1-hot}}$ 
&\textemdash   &5.97$\pm$2.70 &\textemdash &4.65$\pm$1.90   \\         
		BART$_{\textrm{1-hot}}$ 
&\textemdash   &4.46$\pm$2.34 &\textemdash &2.37$\pm$1.34   \\   
		R-learner$_{\textrm{1-hot}}$ 
&\textemdash   &2.52$\pm$0.12 &\textemdash &0.95$\pm$0.25   \\  
		X-learner$_{\textrm{1-hot}}$ 
&\textemdash   &2.64$\pm$1.09 &\textemdash &1.08$\pm$0.49   \\
		OrthoRF$_{\textrm{1-hot}}$ 
&\textemdash   &3.74$\pm$2.24 &\textemdash &2.52$\pm$1.86       
      \\\cmidrule(lr){1-1}\cmidrule(lr){2-3}\cmidrule(lr){4-5}
AdaTRANS  &\textbf{3.60$\pm$2.04}  &\textbf{2.46$\pm$1.09}
		&\textbf{1.94$\pm$1.23} &\textbf{0.70$\pm$0.20}
\\
		\bottomrule
	\end{tabular}
	\label{tab:ihdp-appendix}
\end{table}

\noindent\textbf{Data description.} The Infant Health and Development Program (IHDP) \citep{hill2011bayesian} is a randomized study on the impact of specialist visits (the treatment) on the cognitive development of children (the outcome). The dataset consists of 747 records with 25 covariates describing properties of the children and their mothers. The treatment group includes children who received specialist visits and the control group includes children who did not receive. For each unit, a treated and a control outcome are simulated using the numerical schemes provided in the NPCI package \citep{dorie2016npci}, thus allowing us to know the \textit{true} individual treatment effect.
We use $k$-means on the covariates to divide the data into two sets. Each set is then considered as data from a population. By using $k$-means, we made the data of the two populations different, and thus simulated discrepancy among the populations. We choose the first set as data of the target population and the other set is data of the source population.

\noindent\textbf{Results and discussion.} The baselines in this experiment are also trained in two cases: using `1-hot vector' in combining data and stacking the data. Table~\ref{tab:ihdp-appendix} shows that our method outperform the baselines in both of the evaluation metrics. When training with target data only (0-source), the proposed method still outperforms the baselines, this result shows the superiority of the proposed kernel-based method compared to the other models.

\subsection{Additional results: synthetic data}
This section presents additional results on the synthetic data.
\subsubsection{Transfer analysis}
\begin{figure}[!ht]
    \centering
    \rule{8cm}{0.1pt}
    \\[0cm]
    {\small Varying $d_z$ on different values of $\Delta^\mathsf{s}$, i.e., $(d_z, \Delta^\mathsf{s}) \in \{(35, 0.0), (40, 0.5), (45, 1.0), (50, 1.5), (55,2.0)\}$}
    \\[0.15cm]
    \includegraphics[width=1\textwidth]{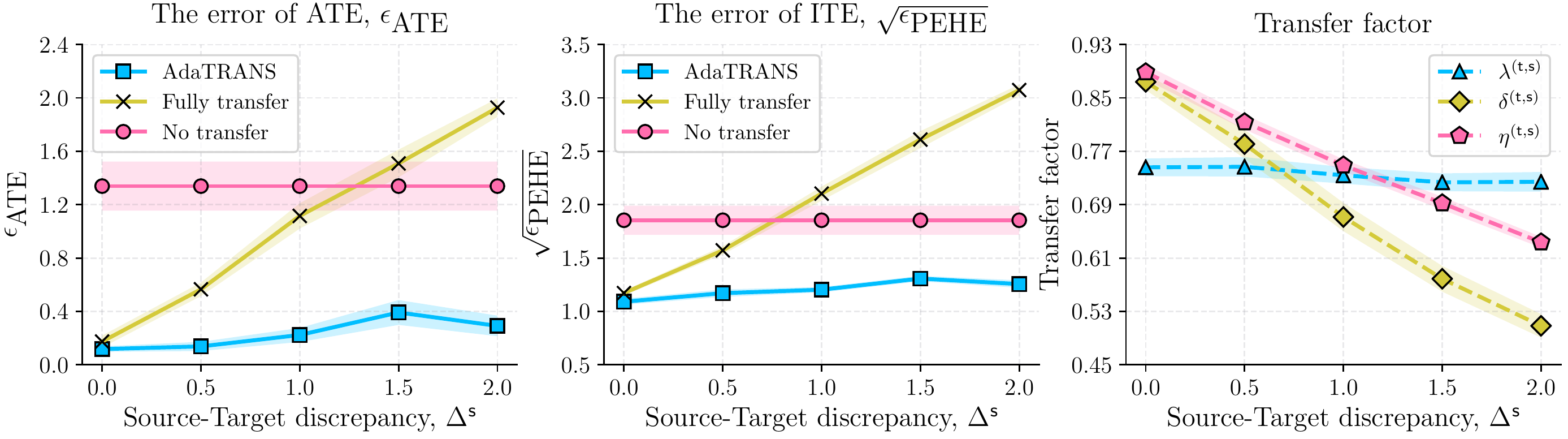}
    \\[0cm]
    \rule{8cm}{0.1pt}
    \\[0cm]
    {\small Fixed $d_z = 40$ for all values of $\Delta^\mathsf{s}$}
    \\[0.15cm]
    \includegraphics[width=1\textwidth]{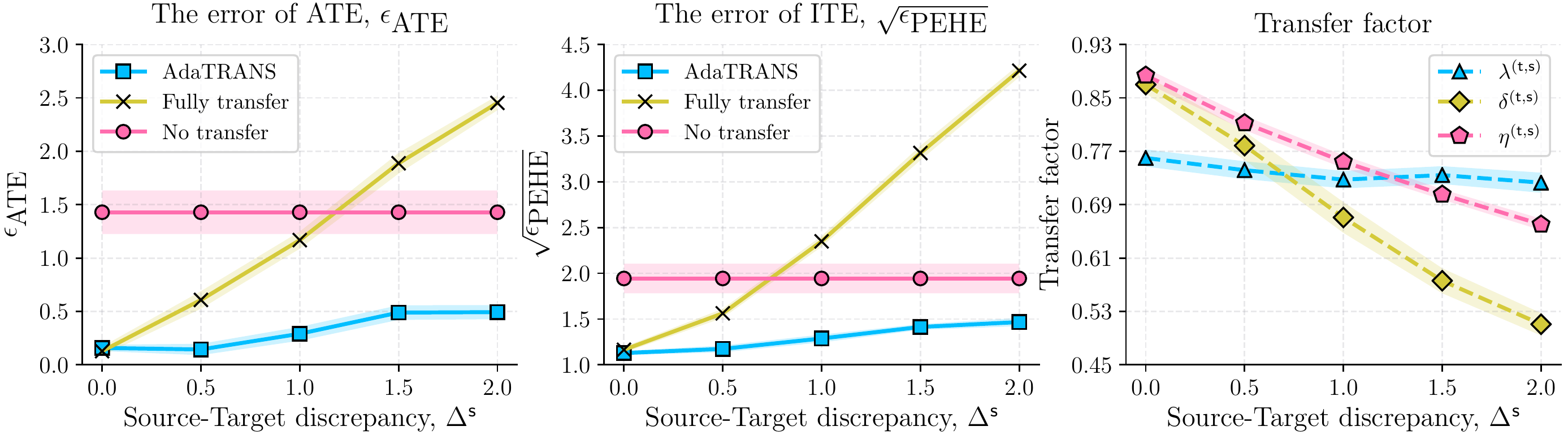}
    \caption{Adaptively causal transfer learning analysis. The figures on the first row are additional results. The figures on the second row were reported in the main text of the paper and repeated here for comparison purpose.}
    \label{fig:adaptively-transfer-analysis-appendix}
\end{figure}

\begin{figure}[!ht]
    \centering
    \includegraphics[width=1\textwidth]{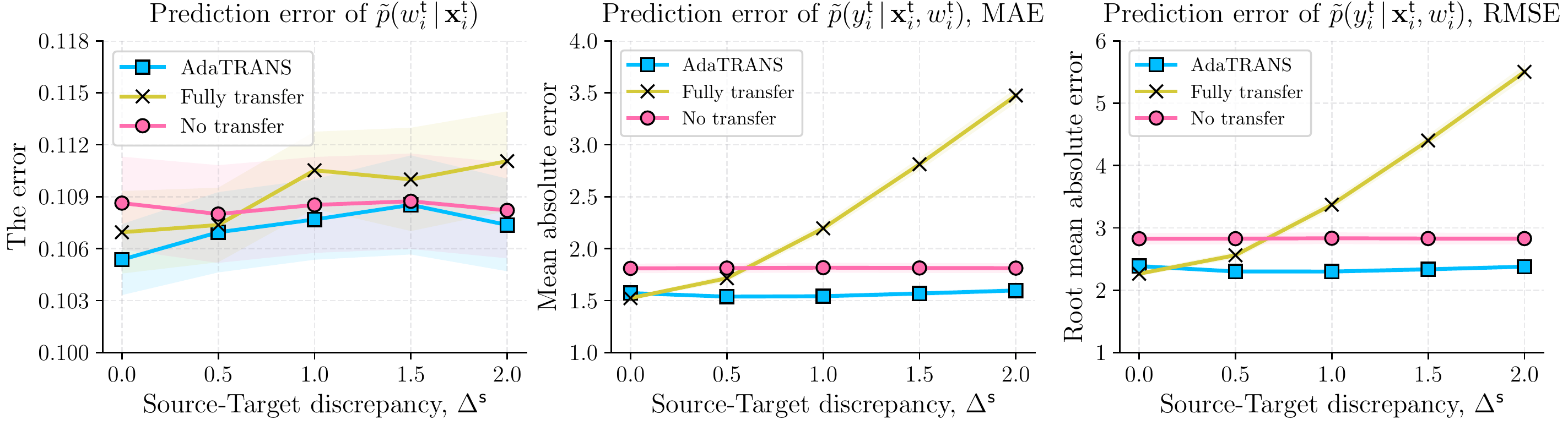}
    \caption{Analysis on prediction error of $w_i^\mathsf{t}$ and $y_i^\mathsf{t}$.}
    \label{fig:prediction-errors-w-y-appendix}
\end{figure}

 Figure~\ref{fig:adaptively-transfer-analysis-appendix} presents the errors in predicting ATE and ITE in two cases: 
\begin{enumerate}[noitemsep,topsep=0pt,leftmargin=*,label=\textbf{(\roman*).}]
    \item Varying the number of dimension of latent confounders (Figure~\ref{fig:adaptively-transfer-analysis-appendix}~(\emph{top})).
    \item Fix the number of dimension of latent confounders (Figure~\ref{fig:adaptively-transfer-analysis-appendix}~(\emph{bottom})). This case is reported in the main text of the paper.
\end{enumerate}
In both cases, the proposed method (adaptively transfer) outperforms the baselines since it only transfer useful information from the sources. It also shows that Case~\textbf{(i)} achieves lower error than that of Case~\textbf{(ii)}. This result indicates that the number of dimensions of latent confounders is important and needs to be tuned in training the model. 
In additional, we also report the prediction errors of $y_i^\mathsf{t}$ and $w_i^\mathsf{t}$ in Figure~\ref{fig:prediction-errors-w-y-appendix}, i.e., they are the errors in learning the predictive distributions $\p(y_i^\mathsf{t}| w_i^\mathsf{t}, \mathbf{x}_i^\mathsf{t})$ (in 2$^{nd}$ transfer level) $\p(w_i^\mathsf{t}| \mathbf{x}_i^\mathsf{t})$ (in 3$^{rd}$ transfer level). Figure~\ref{fig:prediction-errors-w-y-appendix} shows a similar pattern in that of Figure~\ref{fig:adaptively-transfer-analysis-appendix}. This further explains the performance of the proposed model.

\subsubsection{Transfer level analysis} 
\begin{figure}[!ht]
    \centering
    \rule{8cm}{0.1pt}
    \\[0cm]
    {\small Varying $d_z$ on different values of $\Delta^\mathsf{s}$, i.e., $(d_z, \Delta^\mathsf{s}) \in \{(35, 0.0), (40, 0.5), (45, 1.0), (50, 1.5), (55,2.0)\}$}
    \\[0.15cm]
    \includegraphics[width=0.72\textwidth]{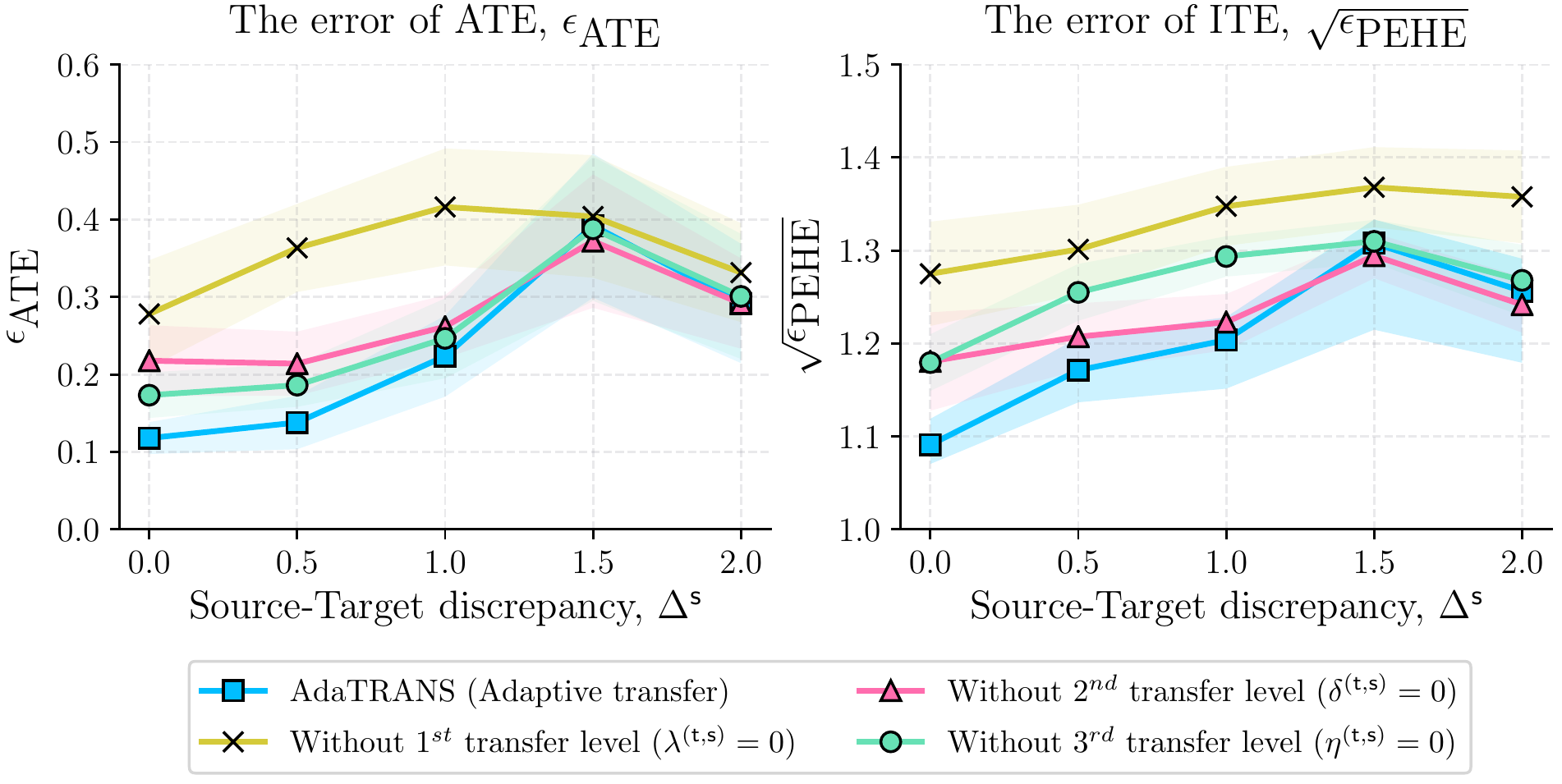}
    \\[0cm]
    \rule{8cm}{0.1pt}
    \\[0cm]
    {\small Fixed $d_z = 40$ for all values of $\Delta^\mathsf{s}$}
    \\[0.15cm]
    \includegraphics[width=0.72\textwidth]{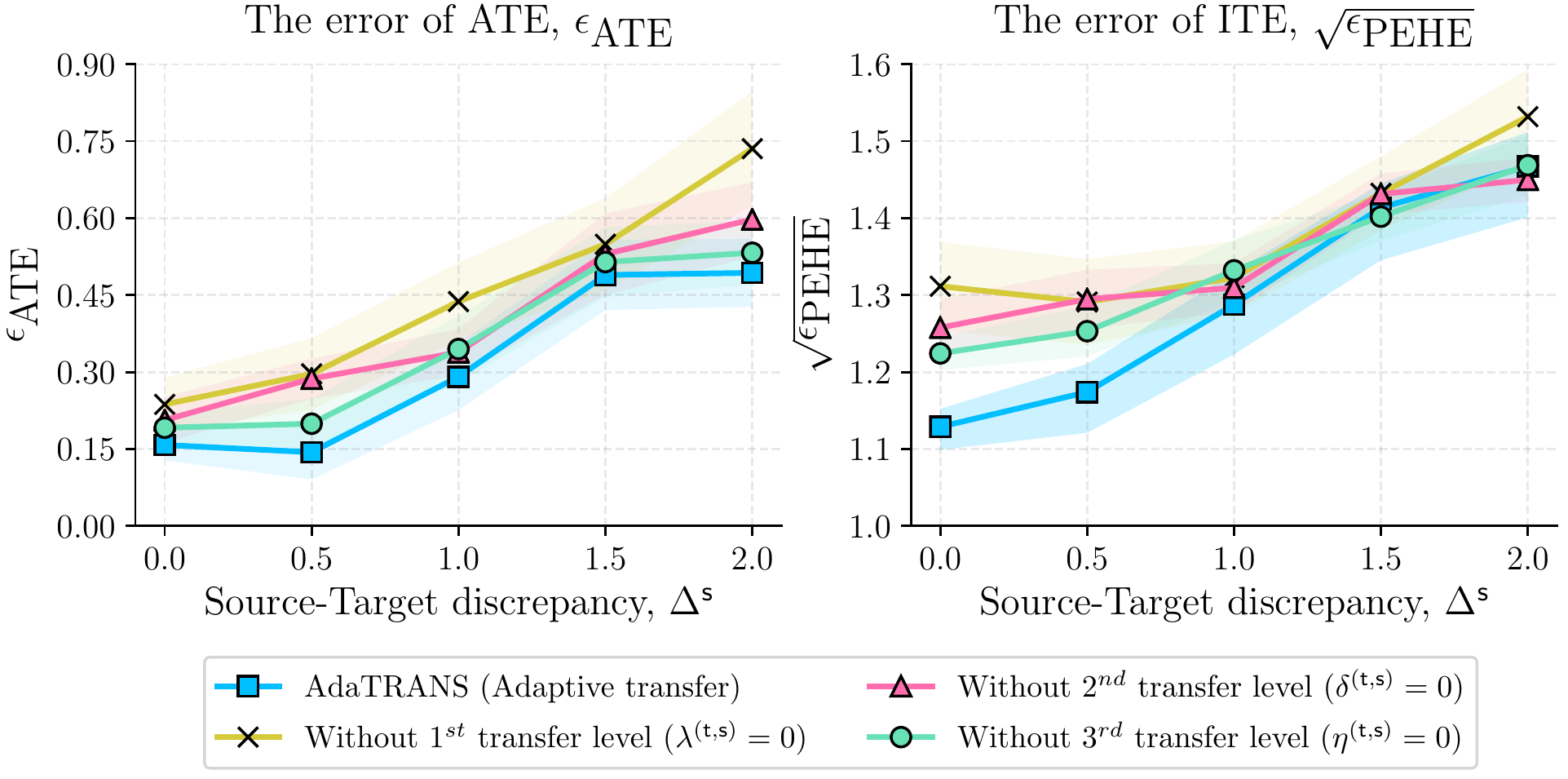}    
    
    \caption{Partially causal transfer learning analysis. The figures on the first row are additional results. The figures on the second row were reported in the main text of the paper and repeated here for comparison purpose.}
    \label{fig:partially-transfer-analysis-appendix}
\end{figure}

Figure~\ref{fig:partially-transfer-analysis-appendix} presents the errors in predicting ATE and ITE when we turn off one level of the knowledge transfer. The figures show that the first level of transfer is the most important since the performance would significantly reduce without this level.

\subsubsection{Prediction error of \texorpdfstring{$\p(w_i^\mathsf{t}|\mathbf{x}_i^\mathsf{t})$}{p(w|x)} and \texorpdfstring{$\p(y_i^\mathsf{t}|w_i^\mathsf{t}, \mathbf{x}_i^\mathsf{t})$}{p(y|wx)} on multi-source data}
\begin{figure}[!ht]
    \centering
    \includegraphics[width=1\textwidth]{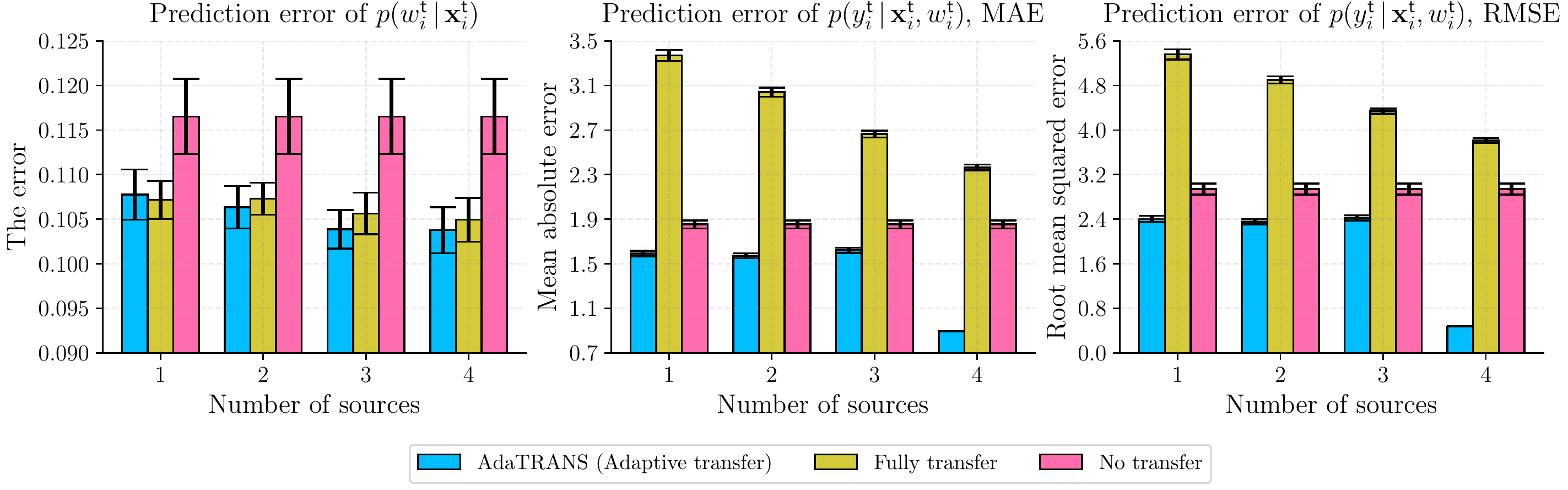}
    \caption{Multi-source analysis on prediction error of $w_i^\mathsf{t}$ and $y_i^\mathsf{t}$.}
    \label{fig:multi-source-analysis-pwx-pywx-appendix}
\end{figure}

Figure~\ref{fig:multi-source-analysis-pwx-pywx-appendix} presents the prediction errors in learning $\p(w_i^\mathsf{t}|\mathbf{x}_i^\mathsf{t})$ and $\p(y_i^\mathsf{t}|w_i^\mathsf{t}, \mathbf{x}_i^\mathsf{t})$ on multiple sources. The figures show that when adding more sources, the errors would decrease as expected. This result further explains the performance in predicting ATE and ITE in Figure~4 of the main text as learning $\p(w_i^\mathsf{t}|\mathbf{x}_i^\mathsf{t})$ and $\p(y_i^\mathsf{t}|w_i^\mathsf{t}, \mathbf{x}_i^\mathsf{t})$ are parts of predicting ATE and ITE.

\subsection{Additional results: Twins dataset}
This section presents details of the specifications used on Twins dataset. We also present comparison of the proposed method with the baseline R-learner (comparison with R-learner was skipped in the main text of the paper for a clear comparison with the other baselines).
\subsubsection{Detailed specifications on Twins dataset}
\begin{figure}
    \centering
    \includegraphics[width=0.7\textwidth]{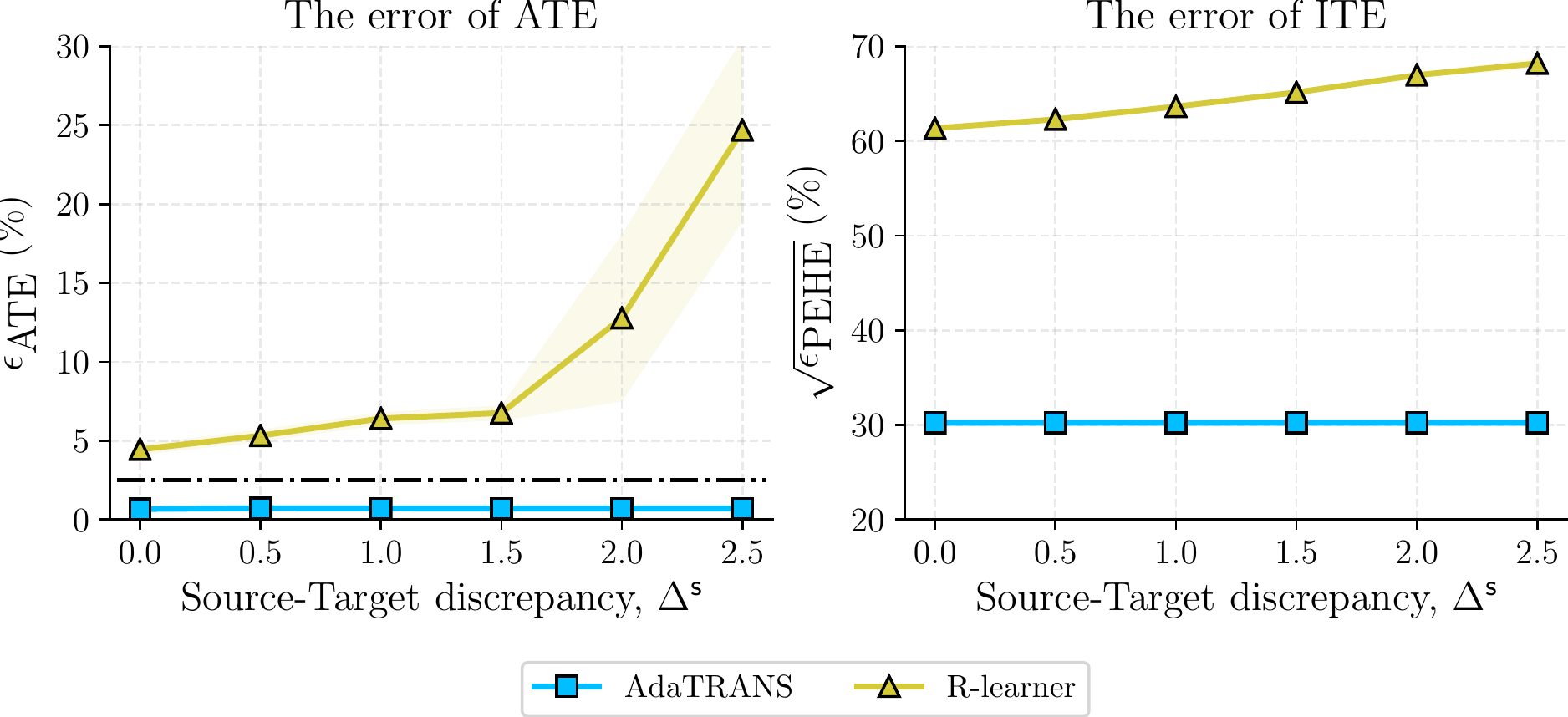}
    \caption{Out-of-sample error of ATE and ITE on Twins dataset.}
    \label{fig:twins-dataset-compare-with-r-learner-appendix-appendix}
\end{figure}
In this section, we present detailed specifications on the Twins dataset. We repeat some descriptions stated in the main manuscript for the completeness.

\noindent\textbf{Data description.} The Twins dataset contains multiple records of twin births in the US from 1989 to 1991 \citep{louizos2017causal}. An abstract treatment $w_i = 1$ corresponds to the twin born with heavier weight and likewise, $w_i = 0$ corresponds to the twin born with lighter weight. The outcome corresponds to the mortality of each of the twins in their first year of life. Since there are records for both twins, the mortality of twins has two possible outcomes (e.g., dead or alive) with respect to the treatment $w_i \in \{0,1\}$. Following \citet{louizos2017causal}, we focused on twins with both weighting less than 2kg.
The observational study is simulated as follows. For each pair of twins, observation regarding one of them is randomly excluded. The entire dataset is then partitioned into two sets: source and target data. The source data accounts for 81\% (3921 entries) and the target data account for 19\% (900 entries). In the target data, we use 9-fold cross-validation with 100 entries for training, 100 for validation, and 700 for testing.

\textbf{Simulation of latent confounders.} To simulate the case of latent confounders with proxy variables, the treatment assignment on twins is based on feature \texttt{GESTAT10}, which records the number of gestation weeks prior to birth and is highly correlated with the mortality outcome. We obtain the observed treatments by drawing from the following distribution $w_i^\mathsf{d}\,|\,z_i^\mathsf{d} \sim \text{Bern}(\varphi(b^\mathsf{d}(0.1z_i^\mathsf{d}-0.1)))$, where $\mathsf{d}\in \{\mathsf{s},\mathsf{t}\}$ and $z_i^\mathsf{d}$ is \texttt{GESTAT10}. We set $b^\mathsf{t} = 0.2$ and $b^\mathsf{s} = b^\mathsf{t} +\Delta^\mathsf{s}$ to simulate discrepancy between the source and target data. We vary $\Delta^\mathsf{s} \in \{0.0, 0.5, 1.0, 1.5, 2.0, 2.5\}$. Following \citet{louizos2017causal}, we created proxies $\mathbf{x}_i^\mathsf{d}$ for the hidden confounder $\mathbf{z}_i^\mathsf{d}$ as follows: The 10 categories of feature \texttt{GESTAT10} are encoded with one-hot encoding, and are replicated three times. We use three replications to ensure that the confounder can be recovered  \citep[follow from][]{kruskal1976more,allman2009identifiability,anandkumar2014tensor,louizos2017causal}. The resulting target data has 30 dimensions for proxy variable $\mathbf{x}_i^\mathsf{d}$.

\subsubsection{Comparison with R-learner on Twins dataset}

Experiment of R-learner was skipped in the main text for a better comparison of the proposed method and the other baselines. Here we report the performance of R-learner compared to the proposed method. In fact, R-learner performs the worst among all the baselines on this dataset. Figure~\ref{fig:twins-dataset-compare-with-r-learner-appendix-appendix} reports the performance of the proposed method (adaptively transfer) compared with R-learner. The figure shows that the proposed method significantly outperforms R-learner on this dataset.

\subsubsection{Additional discussion on Twins dataset}

This section present additional comments on the experimental results. These discussions were skipped in the main text due to limited space.

Here, we further discuss on why the performance in predicting ATE of the other methods are not better than that of the naive estimator. From Figure~5~(left) in the main text, the performance of SITE and X-learner in predicting ATE ($\epsilon_\text{ATE}$) is low, this might be because those methods only consider observed confounders (through the unconfoundedness assumption) and ignore latent confounders. Although CEVAE can handle latent confouders, it is based on deep neural networks which needs a huge dataset to achieve a good performance. Importantly, since SITE and CEVAE are based on neural networks, their performance depends on the tuning of hyperparameters and the optimization algorithm. Hence, they might lead to a local optima. Our method, on the other hand, models complex non-linear function using kernel functions while obtaining convex or conditionally convex objective functions (as stated in Lemma~2 of the main text).

\section{Proof of Lemma 1}

\begin{proof}
We restate that $f_{y_0}\colon \mathcal{Z} \mapsto \mathcal{F}_{y_0}$, $f_{y_1}\colon \mathcal{Z} \mapsto \mathcal{F}_{y_1}$, $f_{q_0}\colon \mathcal{Y} \times \mathcal{X} \rightarrow \mathcal{F}_z$, $f_{q_1}\colon \mathcal{Y} \times \mathcal{X} \rightarrow \mathcal{F}_z$, $f_w\colon \mathcal{Z} \mapsto \mathcal{F}_w$ and $f_x\colon\mathcal{Z} \mapsto \mathcal{F}_x$. In this work, $\mathcal{F}_{y_0} = \mathbb{R}$, $\mathcal{F}_{y_1} = \mathbb{R}$, $\mathcal{F}_w = \mathbb{R}$, $\mathcal{F}_x = \mathbb{R}^{d_x}$ and $\mathcal{F}_z = \mathbb{R}^{d_z}$.

        We further define $f_x = [f_{x,1},\!...,f_{x,d_x}]$ with $f_{x,d}\colon \mathcal{Z}\mapsto \mathbb{R}$ ($d=1,\dots,d_x$). Similarly,  $f_{q_0} = [f_{q_0,1},\!...,f_{q_0,d_z}]$ with $f_{q_0,d}\colon \mathcal{Y}\times \mathcal{X}\mapsto \mathbb{R}$ ($d=1,\dots,d_z$) and $f_{q_1} = [f_{q_1,1},\!...,f_{q_1,d_z}]$ with $f_{q_1,d}\colon \mathcal{Y}\times \mathcal{X}\mapsto \mathbb{R}$ ($d=1,\dots,d_z$).

		Consider the subspaces $\mathcal{U}_c \subset \mathcal{H}_c$, $(c \in \{y_0, y_1, q_0, q_q, x, w\})$ defined as follows:
		\begin{align*}
		\mathcal{U}_{y_0} &= \texttt{span}\big\{\upkappa_{y_0}(\cdot,\mathbf{z}_i^\mathsf{d}[l])\,:\,\mathsf{d} \in \bm{\mathcal{S}}; i=1,\!...,n_\mathsf{d};l=1,\!...,L\big\},\\
		\mathcal{U}_{y_1} &= \texttt{span}\big\{\upkappa_{y_1}(\cdot,\mathbf{z}_i^\mathsf{d}[l])\,:\,\mathsf{d} \in \bm{\mathcal{S}}; i=1,\!...,n_\mathsf{d};l=1,\!...,L\big\},\\
		\mathcal{U}_x &= \texttt{span}\left\{\upkappa_x(\cdot,\mathbf{z}_i^\mathsf{d}[l])\,:\,\mathsf{d} \in \bm{\mathcal{S}};i=1,\!...,n_\mathsf{d};l=1,\!...,L\right\},\\
		\mathcal{U}_w &= \texttt{span}\left\{\upkappa_w(\cdot,\mathbf{z}_i^\mathsf{d}[l])\,:\,\mathsf{d} \in \bm{\mathcal{S}};i=1,\!...,n_\mathsf{d};l=1,\!...,L\right\},\\
		\mathcal{U}_{q_0} &= \texttt{span}\left\{\upkappa_{q_0}(\cdot,[y_i^\mathsf{d},\mathbf{x}_i^\mathsf{d}])\,:\,\mathsf{d} \in \bm{\mathcal{S}};i=1,\!...,n_\mathsf{d}\right\},\\
		\mathcal{U}_{q_1} &= \texttt{span}\left\{\upkappa_{q_1}(\cdot,[y_i^\mathsf{d},\mathbf{x}_i^\mathsf{d}])\,:\,\mathsf{d} \in \bm{\mathcal{S}};i=1,\!...,n_\mathsf{d}\right\}.
		\end{align*} 
		
		We project $f_{y_0}$, $f_{y_1}$, $f_w$, $f_{x,d}$ ($d=1,\!...,d_x$), $f_{q_0,d}$  ($d=1,\!...,d_z$) and $f_{q_1,d}$  ($d=1,\!...,d_z$) onto the subspaces $\mathcal{U}_{y_0}$, $\mathcal{U}_{y_1}$, $\mathcal{U}_w$, $\mathcal{U}_x$, $\mathcal{U}_{q_0}$ and $\mathcal{U}_{q_1}$, respectively, to obtain $f_{y_0}^{\textrm{s}}$, $f_{y_1}^{\textrm{s}}$, $f_w^{\textrm{s}}$, $f_{x,d}^{\textrm{s}}$, $f_{q_0,d}^{\textrm{s}}$ and $f_{q_1,d}^{\textrm{s}}$, and also project them onto the perpendicular spaces of the subspaces to obtain $f_{y_0}^{\bot}$, $f_{y_1}^{\bot}$, $f_w^{\bot}$, $f_{x,d}^{\bot}$, $f_{q_0,d}^{\bot}$ and $f_{q_1,d}^{\bot}$. 
		
		Note that $f_{(\cdot)}^{\textrm{s}} + f_{(\cdot)}^{\bot} = f_{(\cdot)}$. Thus, $\|f_{(\cdot)}\|_{\mathcal{H}_{(\cdot)}}^2 = \|f_{(\cdot)}^{\textrm{s}}\|_{\mathcal{H}_{(\cdot)}}^2 + \|f_{(\cdot)}^{\bot}\|_{\mathcal{H}_{(\cdot)}}^2 \ge \|f_{(\cdot)}^{\textrm{s}}\|_{\mathcal{H}_{(\cdot)}}^2$, which implies that $\lambda_{(\cdot)}\|f_{(\cdot)}\|_{\mathcal{H}_{(\cdot)}}^2$ is minimized if $f_{(\cdot)}$ is in its subspace $\mathcal{U}_{(\cdot)}$. \hfill($\spadesuit$)
		
		Moreover, from the reproducing property, we have that 
\begin{align*}
		f_{y_0}(\mathbf{z}_i^\mathsf{d}[l]) &= \big\langle f_{y_0},\upkappa_{y_0}(\cdot, \mathbf{z}_i^\mathsf{d}[l]) \big\rangle_{\mathcal{H}_y} \\
&= \big\langle f_{y_0}^{\textrm{s}},\upkappa_{y_0}(\cdot, \mathbf{z}_i^\mathsf{d}[l]) \big\rangle_{\mathcal{H}_y} + \big\langle f_{y_0}^{\bot},\upkappa_{y_0}(\cdot, \mathbf{z}_i^\mathsf{d}[l]) \big\rangle_{\mathcal{H}_y}\\
&= f_{y_0}^{\textrm{s}}(\mathbf{z}_i^\mathsf{d}[l]).
		\end{align*}
		Similarly, we have $f_{y_1}( \mathbf{z}_i^\mathsf{d}[l])=f_{y_1}^{\textrm{s}}( \mathbf{z}_i^\mathsf{d}[l])$, $f_w(\mathbf{z}_i^\mathsf{d}[l]) = f_w^{\textrm{s}}(\mathbf{z}_i^\mathsf{d}[l])$, $f_{x,d}(\mathbf{z}_i^l) = f_{x,d}^{\textrm{s}}(\mathbf{z}_i^\mathsf{d}[l])$,  $f_{q_0,d}(y_i^\mathsf{d}, \mathbf{x}_i^\mathsf{d}) = f_{q_0,d}^{\textrm{s}}(y_i^\mathsf{d}, \mathbf{x}_i^\mathsf{d})$ and $f_{q_1,d}(y_i^\mathsf{d}, \mathbf{x}_i^\mathsf{d}) = f_{q_1,d}^{\textrm{s}}(y_i^\mathsf{d}, \mathbf{x}_i^\mathsf{d})$. Hence,
		\begin{align*}
		&\widehat{\mathcal{L}}(f_{y_0}, f_{y_1}, f_{q_0}, f_{q_1}, f_x, f_w) = \widehat{\mathcal{L}}(f_{y_0}^{\textrm{s}}, f_{y_1}^{\textrm{s}}, f_{q_0}^{\textrm{s}}, f_{q_1}^{\textrm{s}}, f_x^{\textrm{s}}, f_w^{\textrm{s}}).
		\end{align*}
		The last equation implies that $\widehat{\mathcal{L}}(\cdot)$ depends only on the component of $f_{y_0}$, $f_{y_1}$, $f_w$, $f_{x,d}$, $f_{q_0,d}$, $f_{q_1,d}$ lying in the subspaces $\mathcal{U}_{y_0}$, $\mathcal{U}_{y_1}$, $\mathcal{U}_w$, $\mathcal{U}_x$, $\mathcal{U}_{q_0}$, $\mathcal{U}_{q_1}$, respectively.\hfill($\clubsuit$)
		
		From ($\spadesuit$) and ($\clubsuit$), we obtain that each $f_c$ is the weighted sum of elements in $\mathcal{U}_c$ ($c \!\in \!\{y_0, y_1, q_0, q_1, x, w\}$). This completes the proof.
\end{proof}

  \section{Proof of Lemma 2}
\begin{proof}
From Lemma~1, we see that the objective function $J$ is a combination of several components including $(\bm{\upalpha}^c)^\top \mathbf{C}\bm{\upalpha}^c$, $ \mathbf{c}^\top\bm{\upalpha}^c$, $-\mathbf{c}^\top\log\varphi(\mathbf{D}\bm{\upalpha}^c)$ and $- (\bm{1}-\mathbf{w})^\top\log\varphi(-\mathbf{D}\bm{\upalpha}^c)$, where $c\in\{y_0, y_1, w, x, q_0, q_1\}$, $\mathbf{C}$ is a positive semi-definite matrix, $\mathbf{c}$ is a vector, and $\mathbf{D}$ is a matrix computed by kernel functions. 

For the first and second term, we have
\begin{align*}
    &\nabla_{\bm{\upalpha}^c}^2\Big\{(\bm{\upalpha}^c)^\top \mathbf{C}\bm{\upalpha}^c\Big\} = \mathbf{C} + \mathbf{C}^\top = 2 \mathbf{C} \succeq 0,\\
    &\nabla_{\bm{\upalpha}^c}^2 \Big\{\mathbf{c}^\top\bm{\upalpha}^c\Big\} = \mathbf{0} \succeq 0,
\end{align*}
where `$\succeq 0$' indicates that the matrix is positive semi-definite.

For the third term, we have
\begin{align*}
    \nabla_{\bm{\upalpha}^c} \Big\{-\mathbf{w}^\top\log\varphi(\mathbf{D}\bm{\upalpha}^c)\Big\} &= - (\nabla_{\bm{\upalpha}^c}\log\varphi(\mathbf{D}\bm{\upalpha}^c))\mathbf{w}\\
&=-(\mathbf{D})^\top\textrm{diag}(\mathbf{w})\varphi(-\mathbf{D}\bm{\upalpha}^c),
\end{align*}
and thus
\begin{align*}
    \nabla_{\bm{\upalpha}^c}^2 \Big\{-\mathbf{w}^\top\log\varphi(\mathbf{D}\bm{\upalpha}^c)\Big\} &= -(\nabla_{\bm{\upalpha}^c}\varphi(-\mathbf{D}\bm{\upalpha}^c))((\mathbf{D})^\top\textrm{diag}(\mathbf{w}))^\top\\
    &= -(\nabla_{\bm{\upalpha}^c}\varphi(-\mathbf{D}\bm{\upalpha}^c))\textrm{diag}(\mathbf{w})\mathbf{D}\\
&=(\mathbf{D})^\top\textrm{diag}(\varphi(-\mathbf{D}\bm{\upalpha}^c)\odot\varphi(\mathbf{D}\bm{\upalpha}^c)\odot \mathbf{w})\mathbf{D} \succeq 0.
\end{align*}
Similarly, for the last term, we have
\begin{align*}
    \nabla_{\bm{\upalpha}^c} \Big\{- (\bm{1}-\mathbf{w})^\top\log\varphi(-\mathbf{D}\bm{\upalpha}^c)\Big\} &=(\mathbf{D})^\top\textrm{diag}(\bm{1}-\mathbf{w})\varphi(\mathbf{D}\bm{\upalpha}^c),
\end{align*}
and
\begin{align*}
       \nabla_{\bm{\upalpha}^c}^2 \Big\{- (\bm{1}-\mathbf{w})^\top\log\varphi(-\mathbf{D}\bm{\upalpha}^c)\Big\} &=(\mathbf{D})^\top\textrm{diag}(\varphi(\mathbf{D}\bm{\upalpha}^c)\odot\varphi(-\mathbf{D}\bm{\upalpha}^c)\odot (\bm{1}-\mathbf{w}))\mathbf{D} \succeq 0.
\end{align*}
Consequently, $J$ is convex because it is a linear combination of convex functions. 
\end{proof}

\end{document}